\documentclass[sigconf]{acmart}
\renewcommand\footnotetextcopyrightpermission[1]{} 

\usepackage{dsfont}

\usepackage{amssymb}
\usepackage{algpseudocode}

\usepackage{soul}
\usepackage{url}
\usepackage[utf8]{inputenc}
\usepackage{caption}
\usepackage{graphicx}
\usepackage{amsmath}
\usepackage{amsthm}
\usepackage{booktabs}
\usepackage{algorithm}
\usepackage{bm}
\usepackage[most]{tcolorbox}

\usepackage{algorithm}
\usepackage{algpseudocode}
\usepackage[switch]{lineno}
\usepackage{changepage}

\usepackage{graphicx}
\usepackage{xcolor}
\definecolor{YouBlue}{RGB}{0, 170, 255}
\definecolor{forestgreen}{RGB}{34,139,34}
\usepackage{fontawesome}

\newcommand{\atk}{\faUserSecret\ }
\newcommand{\defn}{\faShield\ }

\newcommand{\drop}[1]{$_{\textcolor{red}{\downarrow\,#1}}$}

\usepackage{enumitem}
\usepackage{multirow}
\usepackage{subcaption}
\usepackage{tabularx}
\usepackage{natbib}
\usepackage{makecell}
\usepackage{pifont}
\usepackage{fontawesome}
\usepackage{threeparttable}
\usepackage{tikz}

\usepackage{siunitx}
\newtheorem{proposition}{Proposition}
\newtheorem{definition}{Definition}
\newtheorem{lemma}{Lemma}
\newtheorem{theorem}{Theorem}
\newtheorem{assumption}{Assumption}
\newtheorem{problem}{Problem}
\newtheorem{remark}{Remark}
\newtheorem{observation}{Observation}

\newcommand{\nop}[1]{}

\makeatletter
\newcommand{\multiline}[1]{%
  \begin{tabularx}{\dimexpr\linewidth-\ALG@thistlm}[t]{@{}X@{}}
    #1
  \end{tabularx}
}
\makeatother

\AtBeginDocument{%
  }

\setcopyright{acmlicensed}
\copyrightyear{2025}
\acmYear{2025}
\acmDOI{XXXXXXX.XXXXXXX}

\setcopyright{acmlicensed}\acmConference[KDD '27]{33nd ACM SIGKDD Conference on Knowledge Discovery and Data Mining}{August 1-5, 2027}{San Jose, United States}

\acmISBN{978-1-4503-XXXX-X/18/06}

\acmSubmissionID{513}

\begin{document}

\title{BadGCL: Towards Persistent Backdoor Attack on Graph Continual Learning Systems}
\newcommand{\sysname}{BadGCL}

\author{Dong Li}
\affiliation{%
  \institution{Baylor University}
  \city{Waco, Texas}
  \country{USA}
}
\email{dong_li1@baylor.edu}

\author{Yufei Han}
\affiliation{%
  \institution{INRIA}
  \city{Rennes, Ile-de-Vilaine}
  \country{France}
}
\email{yufei.han@inria.fr}

\author{Xintao Wu}
\affiliation{%
  \institution{University of Arkansas}
  \city{Fayetteville, Arkansas}
  \country{USA}
}
\email{xintaowu@uark.edu}

\author{Xujiang Zhao}
\affiliation{%
  \institution{Meta}
  \city{Menlo Park, California}
  \country{USA}
}
\email{zxj32@meta.com}

\author{Wei Cheng}
\affiliation{%
  \institution{NEC Laboratories America}
  \city{Princeton, New Jersey}
  \country{USA}
}
\email{weicheng@nec-labs.com}

\author{Chen Zhao}
\authornote{Corresponding author}
\affiliation{%
  \institution{Baylor University}
  \city{Waco, Texas}
  \country{USA}}
\email{chen_zhao@baylor.edu}

\renewcommand{\shortauthors}{Li et al.}

\begin{abstract}
Backdoor attacks pose a persistent security threat to machine learning systems, yet their behavior under graph continual learning (GCL) remains underexplored.
Conventional graph backdoor attacks often lose effectiveness as models are updated across successive tasks, creating the impression that continual updates naturally weaken malicious associations.
This work reveals a contrasting vulnerability: knowledge-preservation mechanisms in GCL can also preserve early injected backdoors when the poisoned update is compatible with the clean-task signal being retained.
For graph data, this risk is amplified by structural triggers, which are embedded into the message-passing paths of graph neural networks (GNNs) and can induce stable representation shifts across tasks.

To expose this risk, we propose \sysname{}, a model-agnostic framework for persistent backdoor attacks under a strict early-stage intervention setting.
\sysname{} learns topology-aware structural triggers using a gradient mutual-information objective that couples poisoned gradients with clean-task optimization signals, together with a propagation consistency objective that stabilizes trigger-induced effects through graph message passing.
We provide theoretical analysis showing that gradient compatibility and propagation stability jointly support long-term backdoor persistence under benign continual updates.
Extensive experiments across task-IL and class-IL protocols, four GCL benchmarks, and diverse victim systems show that \sysname{} achieves persistent attack effectiveness with limited clean-utility degradation, substantially outperforming graph and continual-learning backdoor baselines and remaining robust against representative defenses.

\end{abstract}

%






\maketitle

\section{Introduction}
\label{sec:introduction}




Graph backdoor attacks have emerged as a critical threat to graph learning systems~\cite{graph_survey,tian2025mldgg}. 
By poisoning a small portion of training graphs with designed structural or feature-based triggers, an attacker can implant a hidden trigger--target mapping into a graph neural network (GNN)~\cite{kipf2016semi,ma2026llm}, such that the model behaves normally on clean inputs but predicts an attacker-chosen label whenever the trigger pattern appears at test time~\cite{li2022backdoor,gao2020backdoor}. 
Prior work has explored trigger designs and injection strategies, ranging from subgraph-based~\cite{sub_backdoor,motif-backdoor,li2023contrastive} and feature-based triggers~\cite{NFTA,shao2024supervised} to optimization~\cite{GTA}, aiming to enhance stealthiness and attack success.

\begin{figure}
    \centering
    \setlength{\abovecaptionskip}{5pt}
    \includegraphics[width=0.9\linewidth]{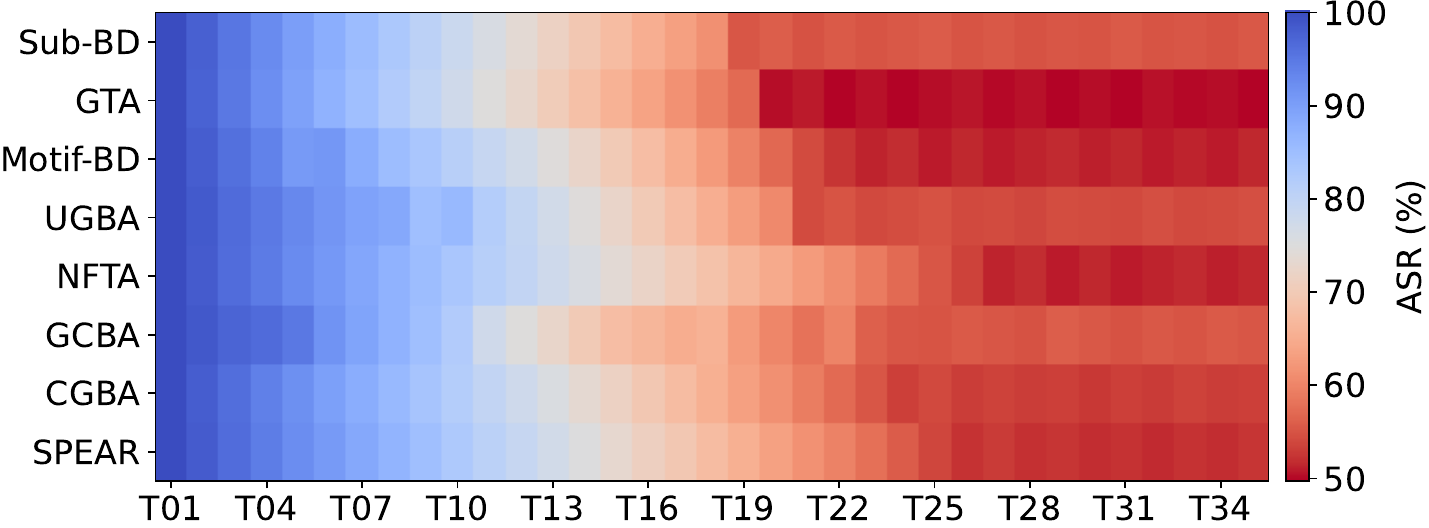}
    \caption{Heatmap of attack success rate (ASR, \%) for eight graph backdoor poisoning methods on \textsc{CoraFull} dataset across 35 tasks, illustrating a progressive decay in attack effectiveness under continual learning setting.}
    \label{fig:intro}
    \vspace{-4mm}
\end{figure}

While traditional backdoor research has predominantly focused on static training and one-shot deployment~\cite{zhang2024rethinking,li2026solverllm,xu2021explainability}, many real-world systems are increasingly dynamic, requiring models to learn from evolving data streams~\cite{latent,CBACL}.
This shift has catalyzed interest in backdoors within continual learning (CL), where models are trained on a sequence of tasks and explicitly optimized to retain historical knowledge~\cite{wang2024comprehensive,li2026hierflow}.
A critical question therefore arises: \emph{how do backdoors behave as models undergo continuous updates?}
Recent studies suggest that conventional backdoors are often non-persistent in CL~\cite{latent,zhao2025uncertainty}.
This trend extends to GCL; as illustrated in Figure~\ref{fig:intro}, primitive poisoning techniques suffer a rapid decay in effectiveness as tasks accumulate.
Importantly, this phenomenon is largely driven by two key bottlenecks of existing graph backdoor attacks in GCL.
First, most trigger designs and injection strategies lack explicit optimization for persistence~\cite{sub_backdoor,li2026marlin}: the trigger--target association often relies on transient correlations that are easily disrupted as the model updates to accommodate new tasks.
Second, existing methods~\cite{GTA,motif-backdoor,CGBA} often struggle to preserve clean utility in GCL: forcing stronger attack effectiveness typically interferes with clean-task retention, making the attack less stealthy and less likely to pass basic performance checks in practical systems.

Together, these bottlenecks create a \emph{false sense of security}: continual adaptation appears to naturally mitigate graph backdoor threats.
Simply adapting existing CL-oriented backdoor designs is also insufficient for GCL because graph triggers are structural: they are embedded into the message-passing paths of GNNs~\cite{xi2021graph,motif-backdoor,tian2026class}, so their persistence depends not only on continual parameter updates but also on how the trigger-induced signal propagates through the graph.
This apparent safety, however, is not a general guarantee.
When the poisoned gradient becomes compatible with the clean-task signal preserved by GCL, and the trigger effect remains stable under message passing, later benign updates can inadvertently retain the malicious association rather than erase it.
In this way, GCL turns its strength in preserving knowledge into a security risk.

To expose and characterize this risk, we propose \sysname{}, a model-agnostic framework for persistent backdoor attacks in GCL under a strict early-stage intervention setting, where the adversary can only poison early training data and has no control over later updates.
The attacker injects graph triggers into a small fraction of early-task samples, after which the victim continues benign GCL; the goal is to preserve attack effectiveness throughout later tasks while maintaining clean utility.
To achieve this, \sysname{} uses the early task to obtain a clean reference optimization signal and learns topology-aware structural triggers with a backdoor objective augmented by a gradient mutual-information objective.
This objective couples the poisoned gradient with the clean-task signal preserved by GCL, making future updates that protect historical knowledge less likely to erase the backdoor.
Meanwhile, a propagation consistency objective stabilizes the trigger-induced representation shift across message-passing depths, and a continuous trigger-topology parameterization with structure regularization learns the trigger connectivity while controlling its structural footprint.

In doing so, \sysname{} addresses both bottlenecks above.
The gradient mutual-information objective reduces backdoor forgetting by making the poisoned update compatible with GCL preservation dynamics, while the propagation objective keeps the trigger effect stable as it travels through graph message passing.
The structure regularizer further improves stealthiness and robustness by avoiding fragile or overly irregular trigger topology.
Together, these components allow \sysname{} to maintain persistent attack effectiveness under benign continual updates with limited clean-utility degradation.
Our main contributions are summarized as follows:
\begin{itemize}[leftmargin=*]
    \item To the best of our knowledge, we present the first effort to expose persistent backdoor threats in GCL systems, revealing a vulnerability: knowledge-preservation mechanisms can be exploited to sustain long-lived backdoor associations when poisoned updates become compatible with preserved clean-task signals.

    \item We propose \sysname{}, a practical framework that learns topology-aware structural triggers through gradient mutual-information optimization, propagation consistency, and structure regularization, enabling persistent attacks without repeated poisoning.

    \item We provide a theoretical characterization of the persistence mechanism, showing how gradient compatibility reduces the initial clean--backdoor loss gap and how propagation stability and structural control bound its drift under benign continual updates.

    \item Through extensive experiments on four GCL benchmarks, we show that \sysname{} outperforms graph and CL backdoor baselines while preserving clean utility. We further validate its robustness under task-order changes, task revisiting, hyperparameter variations, and representative graph-backdoor defenses.
\end{itemize}




\section{Related Work}
\label{sec:relatedwork}
\noindent \textbf{Backdoor Attacks in Graph Mining.}
Graph backdoor attacks have evolved from fixed pattern injection (e.g., subgraph-based attacks~\cite{sub_backdoor} and NFTA~\cite{NFTA}) to optimization-based designs such as GTA~\cite{GTA}, which generates adaptive, input-tailored triggers.
To evade topological detection and structural pruning, UGBA~\cite{UGBA} and SPEAR~\cite{SPEAR} further improve stealthiness via unnoticeability and structure-preserving constraints.
Beyond supervised node classification, backdoors have also been studied in more restrictive settings, including clean-label attacks (CGBA~\cite{CGBA}) and self-supervised graph contrastive learning (GCBA~\cite{GCBA}), with Motif-Backdoor~\cite{motif-backdoor} linking attack efficacy to the statistical significance of injected motifs.
However, these attacks largely assume static training and overlook backdoor forgetting under sequential updates, making them prone to being overwritten in GCL.
In contrast, \sysname{} targets the continual graph setting by learning structural triggers whose poisoned gradients are compatible with clean-task optimization while preserving their propagation-induced effects through message passing, enabling persistent attack effectiveness under GCL updates.

\noindent \textbf{Backdoor Attacks on Continual Learning.}
Recent work has highlighted the vulnerability of CL systems to adversarial poisoning, with a primary focus on backdoor forgetting.
Guo et al.~\cite{latent} pioneered persistent strategies such as Blind Task Backdoor (BTB), which manipulates training loss signals, and Latent Task Backdoor (LTB), which achieves one-shot persistence by anchoring malicious knowledge in the latent representation space.
Attack-on-Prompt~\cite{nguyen2024backdoor} studies prompt-based CL but relies on prompt-tuning mechanisms and is thus incompatible with the message-passing paradigm of GNNs, while CBACL~\cite{CBACL} improves attack flexibility under a more relaxed threat model that allows repeated intervention across tasks.
Despite these advances, existing methods are largely tailored to Euclidean data and struggle to maintain persistence under the restrictive early-stage poisoning setting in GCL.
In contrast, \sysname{} couples the poisoned update with clean-task optimization dynamics and further regularizes the trigger's graph-propagation behavior, allowing the backdoor to persist under replay- and stability-driven GCL without further adversarial intervention.

\section{Preliminaries}
\label{sec:preliminaries}
\label{sec:preliminaries}
\subsection{Victim System}
\label{sec:victim-system}

\noindent \textbf{Graph continual learning.}
We consider a GCL system for node classification that trains a GNN over a sequence of $T$ tasks, denoted as $\{\mathcal{T}_t\}_{t=1}^{T}$.
At each task $t \in \{1, \ldots, T\}$, the learner receives a labeled training dataset $\mathcal{D}_t^{\mathrm{tr}} = \{(x_i, y_i)\}_{i=1}^{n_t}$ and is evaluated on a corresponding test set $\mathcal{D}_t^{\mathrm{te}}$.
Each instance $x$ represents a target node along with its $k$-hop neighborhood subgraph, associated with a label $y \in \mathcal{Y}_t$, where $\mathcal{Y}_t$ denotes the label space for task $t$.
Let $f_{\theta_t}: \mathcal{X} \to \mathbb{R}^{|\mathcal{Y}_{\le t}|}$ denote the $K$-layer GNN parameterized by $\theta_t$, which maps a graph input to a logit vector over the cumulative label space $\mathcal{Y}_{\le t} = \bigcup_{k=1}^{t}\mathcal{Y}_k$.
The system updates model parameters sequentially, yielding a parameter trajectory $\{\theta_t\}_{t=1}^{T}$, where $\theta_t$ represents the model parameters after training on tasks $1$ through $t$.
The predicted label for input $x$ is obtained via:
\begin{equation*}
\hat{y}_{\theta_t}(x) = \arg\max_{y\in \Omega_t} \; [f_{\theta_t}(x)]_y ,
\label{eq:pred}
\end{equation*}
where $[f_{\theta_t}(x)]_y$ denotes the logit corresponding to label $y$, and $\Omega_t \subseteq \mathcal{Y}_{\le t}$ is the candidate label set determined by the GCL protocol.

\noindent \textbf{Continual learning protocols.}
We consider two standard GCL protocols that differ in the information available at inference time: \textit{(i) Task-Incremental Learning (Task-IL).}
The task identity $t$ is provided at test time, so the prediction is restricted to the label set of the current task, i.e., $\Omega_t^{\mathrm{Task\text{-}IL}} = \mathcal{Y}_t$. \textit{(ii) Class-Incremental Learning (Class-IL).}
The task identity is unavailable at test time, requiring the model to discriminate among all labels observed so far, i.e., $\Omega_t^{\mathrm{Class\text{-}IL}} = \mathcal{Y}_{\le t}$.
Class-IL is significantly more challenging than Task-IL due to the need to distinguish between classes across all tasks without explicit task boundaries~\cite{survey}.


\subsection{Threat Model}
\label{sec:threat-model}

\noindent \textbf{Attacker's knowledge.}
The attacker has white-box access to the victim system during the intervention phase $t< t'$, including the GNN architecture $f_\theta$, the CL algorithm, and the deployment protocol.
However, the attacker has no knowledge of future task distributions $\{\mathcal{D}_t\}_{t \ge t'}$ or label spaces $\{\mathcal{Y}_t\}_{t \ge t'}$.

\noindent \textbf{Attacker's capability.}
The attacker can intervene only during an \emph{early stage} of CL, with no access to subsequent tasks.
In this work, we focus on the constrained scenario where intervention is limited to the first task $\mathcal{T}_1$.
Within this window, the attacker can arbitrarily manipulate the training process (loss functions, optimization, parameter updates) and poison up to an $\epsilon$-fraction of $\mathcal{D}_1^{\mathrm{tr}}$ where $\epsilon \ll 1$.
All subsequent training proceeds benignly without adversarial interference.

\noindent \textbf{Attack goal.}
Let $G_{\Delta}$ denote a graph trigger pattern and $x_{\Delta} = \mathcal{I}(x,G_{\Delta})$ denote the triggered version of a clean input graph $x$, where $\mathcal{I}(\cdot,\cdot)$ is an injection operator.
The attacker selects a target label $y^\star$ and aims to achieve three objectives: (i) targeted misclassification on triggered graph inputs throughout the CL trajectory, (ii) utility preservation on clean inputs during the whole process, and (iii) temporal persistence of the backdoor poisoning effects despite subsequent benign updates on tasks $\{\mathcal{T}_2, \ldots, \mathcal{T}_T\}$.


\subsection{Problem Formulation}
\label{sec:problem-formulation}

We formulate persistent backdoor poisoning in GCL as an early-stage constrained optimization problem.
Unlike static backdoor attacks, which only need to balance clean utility and attack effectiveness on a fixed training distribution, a GCL backdoor must also remain effective after a sequence of future benign updates.
Thus, the trigger must not only induce the target-label association during the intervention window, but also preserve this association along the continual learning trajectory.

\begin{problem}[Persistent Backdoor Attack on Graph Continual Learning]
\label{prob:persistent-backdoor}
Given a graph continual learning system with task sequence $\{\mathcal{T}_t\}_{t=1}^{T}$, an early intervention window ending at task $t'$, and a target label $y^\star$, the attacker can poison only the training data within the intervention window.
Let $\mathcal{A}\subseteq \mathcal{D}_{\le t'}^{\mathrm{tr}}$ be the attacked subset with $|\mathcal{A}|\le \epsilon|\mathcal{D}_{\le t'}^{\mathrm{tr}}|$, where $\epsilon\ll1$, and let
\[
\mathcal{D}^{\mathrm{bd}}
=
\{(\mathcal{I}(x,G_\Delta),y^\star)\mid (x,y)\in\mathcal{A}\}
\]
be the poisoned samples generated by trigger $G_\Delta$.
The attacker seeks a trigger $G_\Delta^\star$ and a poisoned checkpoint $\theta_{t'}^\star$ by solving
\begin{equation*}
\min_{G_\Delta,\theta_{t'}}
\;
\mathcal{L}_{\mathrm{cl}}(\theta_{t'};\mathcal{D}_{\le t'}^{\mathrm{tr}}\setminus\mathcal{A})
+
\lambda\mathcal{L}_{\mathrm{bd}}(\theta_{t'};\mathcal{D}^{\mathrm{bd}}),
\label{eq:problem-obj}
\end{equation*}
subject to the temporal persistence constraints after benign future updates:
\begin{equation*}
S_t(\theta_t;G_\Delta^\star)
\ge
S_{t'}(\theta_{t'}^\star;G_\Delta^\star)-\delta_{\mathrm{attack}},
\quad
U_t(\theta_t)
\ge
U_t^{\mathrm{ben}}-\delta_{\mathrm{cl}},
\quad
\forall t>t',
\label{eq:problem-constraints}
\end{equation*}
where subsequent parameters are obtained only through benign GCL updates on future tasks.
Here,
\begin{equation*}
S_t(\theta_t;G_\Delta^\star)
\triangleq
\mathbb{P}_{(x,y)\sim\mathcal{D}^{\mathrm{bd,te}}}
\!\left[
\hat{y}_{\theta_t}(\mathcal{I}(x,G_\Delta^\star))=y^\star
\right],
\label{eq:problem-asr}
\end{equation*}
and $U_t(\theta_t)$ denotes the backdoor attack success probability and the clean utility of the GCL system after learning tasks up to $t$ respectively under the corresponding evaluation protocol.
$U_t^{\mathrm{ben}}$ denotes the clean utility of the poisoning-free victim system under the same protocol, while $\delta_{\mathrm{attack}},\delta_{\mathrm{cl}}\ge0$ specify the tolerated degradation in attack effectiveness and clean utility.
\end{problem}

The first constraint enforces \emph{backdoor persistence}, requiring the ASR on triggered test inputs to remain stable after the attacker exits.
The second enforces \emph{utility preservance}, requiring the poisoned model to maintain clean utility close to the benign victim system.
In this work, we instantiate the attack with $t'=1$, where the attacker can intervene only on the first task.

\section{Security Vulnerability of GCL System}
\label{sec:vulnerability}
As illustrated in Figure~\ref{fig:intro}, one may expect backdoors implanted during early tasks of a GCL system to naturally decay as the model undergoes sequential benign updates without further adversarial intervention. This is due to \emph{catastrophic forgetting} of DNN models~\cite{goodfellow2013empirical}. This difficulty renders the preservation of backdoor efficacy in Problem~\ref{prob:persistent-backdoor} a formidable task.
However, we identify a \emph{security vulnerability} in GCL systems:

\begin{observation}[The Persistence Paradox]
\label{obs:persistence_in_gcl}
    Anti-forgetting mechanisms in GCL, while essential for task retention, can be inadvertently hijacked to consolidate and stabilize structural backdoors.
\end{observation}

The key reason is that a graph backdoor can be made to behave like a protected structural signal rather than a transient input perturbation.
A subgraph trigger rewires the local message-passing computation graph and induces a propagation-dependent representation shift whenever it is present.
If this shift is further optimized to be compatible with the clean-task optimization signal preserved by GCL, later benign updates that retain historical knowledge may also retain the trigger-induced effect.
Thus, the backdoor can ``free ride'' on anti-forgetting: the same stability mechanisms that constrain clean-task degradation can also limit the drift of a well-optimized backdoor.
This persistence is not tied to a specific GCL algorithm, but to the broader objective of preserving task-relevant structural and optimization signals, enabling an early-stage structural attack to remain effective throughout the continual learning trajectory.

\section{Persistent Backdoor Attack on GCL}
\label{sec:method}
\label{sec:method}
 
 
To validate and quantify Observation~\ref{obs:persistence_in_gcl}, we propose \sysname{}, a persistent backdoor attack framework for GCL under the early-stage intervention constraint in Problem~\ref{prob:persistent-backdoor}. The key idea is to learn a structural trigger that is both \emph{optimization-compatible} with the clean-task knowledge preserved by GCL and \emph{propagation-stable} under GNN message passing. During the attacker's only intervention window on task $\mathcal{T}_1$, \sysname{} first trains a clean reference model to characterize the initial task's optimization signal, and then optimizes a subgraph trigger with two complementary objectives: \emph{a gradient mutual-information based objective} that couples poisoned updates with clean-task optimization, and \emph{a propagation consistency objective} that stabilizes the trigger-induced effect across message-passing steps. The learned trigger is then implanted through poisoning fine-tuning while preserving clean utility. After this one-shot early poisoning, the attacker exits permanently and the victim continues standard benign continual learning on subsequent tasks. 

 
 
 
\subsection{Learnable Topology-Aware Triggers}
\label{sec:trigger_design}
 
We adopt a subgraph trigger that perturbs the local topology of input graphs.
Let $x$ denote an input graph instance for node classification with $d$-dimensional node features.
The trigger is a small graph $G_{\Delta}=(X_{\Delta},A_{\Delta})$ with $m$ nodes, where $m$ is much smaller than the size of the target graph, and $X_\Delta\in\mathbb{R}^{m\times d}$ contains learnable trigger-node features.
 
\noindent \textbf{Learnable trigger topology.}
We parameterize the trigger topology with learnable edge logits $A_{\Delta}\in\mathbb{R}^{m\times m}$, initialized with a high edge probability for reliable activation, and control its density via
\begin{equation}
\mathcal{R}_{\mathrm{str}}(A_{\Delta})
= \left( \frac{1}{m^2} \mathbf{1}^\top \sigma(A_{\Delta}) \mathbf{1} - \rho \right)^2,
\label{eq:struct-reg}
\end{equation}
where $\sigma(\cdot)$ is the sigmoid function and $\rho\in(0,1)$ is the desired density.
During optimization, $\sigma(A_{\Delta})$ serves as the soft trigger adjacency; during deployment, we discretize it by thresholding $\hat{A}_{\Delta}=\mathbb{I}(\sigma(A_{\Delta})>\tau)$ with $\tau=0.5$.
 
\noindent \textbf{Poisoned data generation.}
Given $\mathcal{D}_1^{\mathrm{tr}}$, we sample an $\epsilon$-fraction of training instances as anchors $\mathcal{A}\subset \mathcal{D}^{\mathrm{tr}}_1$ and construct
\begin{equation}
\mathcal{D}^{\mathrm{bd}} \triangleq \{(x_{\Delta}, y^\star)\mid (x,y)\in \mathcal{A}\},
\quad
\mathcal{D}^{\mathrm{poison}} \triangleq (\mathcal{D}_1^{\mathrm{tr}}\setminus \mathcal{A}) \cup \mathcal{D}^{\mathrm{bd}},
\label{eq:dataset_split}
\end{equation}
where $x_{\Delta}=\mathcal{I}(x,G_{\Delta})$ is the triggered input and $y^\star$ is the target label.
The injection operator $\mathcal{I}(\cdot,\cdot)$ augments $x$ with the $m$ trigger nodes, their learned internal topology, and a small attachment between the anchor node and the trigger.
This minimally perturbs the global semantics while allowing the trigger to participate in local message-passing dynamics.
 
\subsection{Compatibility- and Propagation-Aware Poisoning}
\label{sec:poisoning}
 
We now describe the poisoning procedure executed during the attacker's only intervention window on task $\mathcal{T}_1$.
\sysname{} follows a three-phase design: \emph{clean pre-training}, \emph{trigger optimization with frozen model parameters}, and \emph{poisoning fine-tuning}.
 
\noindent \textbf{Phase 1: Clean pre-training.}
We first train the victim GNN $f_\theta$ on the clean first-task data $\mathcal{D}_1^{\mathrm{tr}}$ using the standard supervised loss:
\begin{equation*}
\begin{split}
    \theta_{\mathrm{cl}}
    &= \arg\min_\theta \mathcal{L}_{\mathrm{cl}}(\theta;\mathcal{D}_1^{\mathrm{tr}}),\quad \mathcal{L}_{\mathrm{cl}}(\theta;\mathcal{D})
    = \mathbb{E}_{(x,y)\sim\mathcal{D}}
    \left[\ell(f_\theta(x),y)\right].
\end{split}
\end{equation*}
The resulting model $\theta_{\mathrm{cl}}$ serves as a clean reference state for $\mathcal{T}_1$.
We then compute the clean-task optimization signal
\begin{equation*}
g_{\mathrm{cl}}
= \nabla_\theta \mathcal{L}_{\mathrm{cl}}(\theta;\mathcal{D}_1^{\mathrm{tr}})
\big|_{\theta=\theta_{\mathrm{cl}}},
\end{equation*}
which characterizes the task-relevant descent direction that GCL mechanisms are encouraged to preserve.

\noindent \textbf{Phase 2: Trigger optimization.}
With $\theta_{\mathrm{cl}}$ fixed, we optimize only the trigger $G_{\Delta}=(X_\Delta,A_\Delta)$.
Given the anchor set $\mathcal{A}$ in Eq.~\eqref{eq:dataset_split}, the backdoor objective is
\begin{equation*}
\mathcal{L}_{\mathrm{bd}}(\theta;G_{\Delta},\mathcal{A})
= \mathbb{E}_{(x,y)\sim\mathcal{A}}
\left[
\ell(f_\theta(\mathcal{I}(x,G_{\Delta})),y^\star)
\right],
\end{equation*}
where $\mathcal{I}$ is the trigger insertion operator.
This term makes the trigger activatable under the clean reference model. To ensure the persistence (the trigger survives benign updates),  
we maximize the mutual information (MI) between the poisoned gradient
$g_{\mathrm{bd}}(G_{\Delta})=\nabla_\theta\mathcal{L}_{\mathrm{bd}}(\theta;G_{\Delta},\mathcal{A})|_{\theta=\theta_{\mathrm{cl}}}$
and the clean gradient signal $g_{\mathrm{cl}}$:
\begin{equation*}
I(g_{\mathrm{bd}};\,g_{\mathrm{cl}})
=
\mathbb{E}_{p(g_{\mathrm{bd}},g_{\mathrm{cl}})}
\left[\log\frac{p(g_{\mathrm{bd}},g_{\mathrm{cl}})}
{p(g_{\mathrm{bd}})\,p(g_{\mathrm{cl}})}\right],
\end{equation*}
where the distributions arise from trigger stochasticity and mini-batch sampling.
Maximizing $I(g_{\mathrm{bd}};\,g_{\mathrm{cl}})$ encourages the backdoor
update to systematically co-vary with the task-relevant descent signal
preserved by GCL, rather than merely matching a single sampled gradient
direction.
A natural alternative is to minimize the averaged Euclidean or cosine
distance between the two gradient vectors~\cite{P-Trojan}; however, such
point-wise metrics reduce each gradient pair to a scalar similarity score
and then average over mini-batches, discarding the joint distributional
structure, i.e., \emph{how} the backdoor and clean gradient directions
co-vary across stochastic mini-batch realizations. MI instead quantifies the full statistical dependency between the
backdoor-poisoned and clean gradient distributions, capturing both linear
and nonlinear coupling.
It is further invariant to invertible reparameterizations of the gradient space, making it an intrinsic compatibility measure robust to the non-uniform parameter scaling that arises across GNN layers.

\noindent\textbf{From MI to InfoNCE.}
Exact computation of $I(g_{\mathrm{bd}};g_{\mathrm{cl}})$ requires estimating an intractable density ratio for high-dimensional gradients.
We therefore minimize an InfoNCE contrastive loss~\cite{Oord2018RepresentationLW}, which provides a tractable MI lower bound.
Let $\{g_r^{-}\}_{r=1}^{R}$ be negative clean gradients from randomly sampled non-anchor subsets of $\mathcal{D}_1^{\mathrm{tr}}\setminus\mathcal{A}$.
Using cosine similarity $s(a,b)=\langle a,b\rangle/(\|a\|_2\|b\|_2)$ and temperature $\tau$, define
$q_0=s(g_{\mathrm{bd}}(G_{\Delta}),g_{\mathrm{cl}})$ and
$q_r=s(g_{\mathrm{bd}}(G_{\Delta}),g_r^{-})$ for $r\ge1$.
The gradient MI loss is
\begin{equation*}
\mathcal{L}_{\mathrm{MI}}(G_{\Delta})
=
-\log
\frac{\exp(q_0/\tau)}
{\sum_{r=0}^{R}\exp(q_r/\tau)}\,.
\end{equation*}
Proposition~\ref{prop:infonce-mi} in Appendix~\ref{sec:theory-infonce} establishes
$I(g_{\mathrm{bd}};\,g_{\mathrm{cl}}) \ge \log(R+1)-\mathcal{L}_{\mathrm{MI}}$,
so minimizing $\mathcal{L}_{\mathrm{MI}}$ maximizes a lower bound on MI.
Compared with cosine similarity alone, the $(R{+}1)$-way softmax is stronger: it aligns $g_{\mathrm{bd}}$ with $g_{\mathrm{cl}}$ while separating it from non-matching clean directions $\{g_r^{-}\}$, ensuring that $g_{\mathrm{bd}}$ shares \emph{the clean-task optimization structure} rather than merely following one transient direction (Lemma~\ref{lem:mi-sep} in Appendix~\ref{sec:theory-lemmas}).
The softmax also estimates pointwise mutual information, capturing statistical co-variation beyond any single-pair similarity score.

\noindent\textbf{Propagation consistency.}
Beyond gradient-level compatibility, graph triggers influence the target node through local message passing, so their effect must also be stable across propagation depths.
Let $\mathcal{G}_{\mathcal{A}}$ denote the local computation graph around the poisoned anchors, on which we compare the clean and triggered propagation states.
For an anchor node $v\in\mathcal{A}$, let $h_v^{(k)}$ and $h_{\Delta,v}^{(k)}$ denote its representation after the $k$-th propagation step before and after trigger injection, respectively, and define
\begin{equation*}
r_{\Delta}^{(k)}(v)
=
h_{\Delta,v}^{(k)} - h_v^{(k)} .
\end{equation*}
We encourage this shift to remain stable across propagation depths by minimizing
\begin{equation*}
\begin{aligned}
\mathcal{L}_{\mathrm{prop}}(G_{\Delta})
=
\frac{1}{|\mathcal{A}|(K-1)}
&\sum_{v\in\mathcal{A}}
\sum_{k=2}^{K}
\Big[
d_{\mathrm{dir}}\!\left(r_{\Delta}^{(k)}(v),r_{\Delta}^{(1)}(v)\right) \\
&+
\gamma
\left(
\|r_{\Delta}^{(k)}(v)\|_2-\|r_{\Delta}^{(1)}(v)\|_2
\right)^2
\Big].
\end{aligned}
\end{equation*}
where
$d_{\mathrm{dir}}(a,b)=\left\|\frac{a}{\|a\|_2}-\frac{b}{\|b\|_2}\right\|_2^2$
and $\gamma$ controls the magnitude-consistency term. $K$ is the number of message-passing layers in the GNN model. 
The final trigger optimization objective is
\begin{multline*}
G_{\Delta}^\star
=
\arg\min_{G_{\Delta}}
\;
\mathcal{L}_{\mathrm{bd}}(\theta_{\mathrm{cl}};G_{\Delta},\mathcal{A})
+
\lambda_{\mathrm{MI}}\mathcal{L}_{\mathrm{MI}}(G_{\Delta})
\\
+
\lambda_{\mathrm{prop}}\mathcal{L}_{\mathrm{prop}}(G_{\Delta})
+
\lambda_{\mathrm{s}}\mathcal{R}_{\mathrm{str}}(A_\Delta),
\end{multline*}
where $\lambda_{\mathrm{MI}}$, $\lambda_{\mathrm{prop}}$, and $\lambda_{\mathrm{s}}$ balance gradient compatibility, propagation stability, and structural stealthiness.
 
\noindent \textbf{Phase 3: Poisoning fine-tuning.}
After obtaining $G_{\Delta}^\star$, we fix the trigger and fine-tune the model from $\theta_{\mathrm{cl}}$ to implant the backdoor while preserving clean utility:
\begin{equation*}
\theta_1^\star
=
\arg\min_{\theta}
\;
\mathcal{L}_{\mathrm{cl}}(\theta;\mathcal{D}_1^{\mathrm{tr}}\setminus\mathcal{A})
+
\lambda_{\mathrm{bd}}
\mathcal{L}_{\mathrm{bd}}(\theta;G_{\Delta}^\star,\mathcal{A}),
\ 
\theta^{(0)}=\theta_{\mathrm{cl}}.
\end{equation*}
The fine-tuning phase is kept brief to avoid overfitting to the poisoned anchors.
The resulting state $\theta_1^\star$ encodes both clean first-task knowledge and the trigger-target association.
 
After poisoning fine-tuning, the attacker exits permanently, and the victim continues standard GCL on $\{\mathcal{T}_2,\ldots,\mathcal{T}_T\}$ using only benign data.
Since $G_{\Delta}^\star$ is optimized for both clean-task compatibility and propagation stability, later anti-forgetting updates are less likely to erase the backdoor and may instead retain it together with preserved clean knowledge.
Algorithm~\ref{alg:overall} in Appendix~\ref{sec:app_alg_overall} summarizes the overall procedure.

\subsection{Attack Feasibility Analysis}
\label{sec:feasibility}

We provide a theoretical characterization of the persistence of \sysname{} after the attacker exits and the GNN undergoes benign parameter updates of downstream GCL tasks.
Full supporting lemmas and proofs are provided in Appendix~\ref{sec:theory}.

\noindent\textbf{Setup and central quantity.}
Let $\mathcal{L}_{\mathrm{cl}}(\theta)=\mathbb{E}_{(x,y)\sim\mathcal{D}_1^{\mathrm{tr}}}[\ell(f_\theta(x),y)]$ be the \emph{clean first-task loss} and $\mathcal{L}_{\mathrm{bd}}(\theta)=\mathbb{E}_{(x,y)\sim\mathcal{A}}[\ell(f_\theta(\mathcal{I}(x,G_\Delta)),y^\star)]$ the \emph{backdoor loss}, with corresponding gradients $g_{\mathrm{cl}}(\theta)$ and $g_{\mathrm{bd}}(\theta)$.
We define the \emph{target-task loss gap} as
\begin{equation}\label{eq:gap}
    \Phi(\theta)
    \;=\;
    \mathcal{L}_{\mathrm{bd}}(\theta)
    \;-\;
    \mathcal{L}_{\mathrm{cl}}(\theta),
\end{equation}
We write $L=\max(L_{\mathrm{bd}},L_{\mathrm{cl}})$ for the Lipschitz smoothness constant and $D_t=\max_{1\le i\le t}\|\theta_i-\theta_{\mathrm{cl}}\|_2$ for the trajectory diameter.

\begin{theorem}[MI-Driven Base Gap Minimization]\label{thm:mi-gap}
    Let $g_{\mathrm{cl}}$ and $g_{\mathrm{bd}}$ be nonzero gradients at $\theta_{\mathrm{cl}}$.
    If the MI objective achieves $\mathcal{L}_{\mathrm{MI}}\le\epsilon_{\mathrm{MI}}$ with $\epsilon_{\mathrm{MI}}<\log 2$ and the non-degeneracy condition holds (i.e., at least one negative gradient satisfies $s(g_{\mathrm{bd}}, g_r^-)\ge 0$; see Lemma~\ref{lem:mi-sep}), then $\kappa=s(g_{\mathrm{bd}},g_{\mathrm{cl}})\ge\kappa(\epsilon_{\mathrm{MI}})>0$, and:

    \textbf{(a)} In Phase~3, $\Delta\theta=-\eta(g_{\mathrm{cl}}+\lambda_{\mathrm{bd}}g_{\mathrm{bd}})$
    is a descent direction for \emph{both} $\mathcal{L}_{\mathrm{cl}}$ and $\mathcal{L}_{\mathrm{bd}}$ for any $\lambda_{\mathrm{bd}}>0$ and sufficiently small $\eta>0$.

    \textbf{(b)} After $E$ steps of Phase~3 fine-tuning with sufficiently small $\eta$,
    \begin{equation}\label{eq:thm2}
        \Phi(\theta_1^\star)
        \;\le\;
        \Phi(\theta_{\mathrm{cl}})
        \;-\;
        \eta E\,\Psi(\kappa,\lambda_{\mathrm{bd}})
        \;+\;
        \mathcal{O}(\eta^2 E),
    \end{equation}
    where
    $\Psi(\kappa,\lambda_{\mathrm{bd}})
    =
    \lambda_{\mathrm{bd}}\|g_{\mathrm{bd}}\|_2^2
    -\|g_{\mathrm{cl}}\|_2^2
    +(1-\lambda_{\mathrm{bd}})\kappa\|g_{\mathrm{bd}}\|_2\|g_{\mathrm{cl}}\|_2$
    is the per-step gap reduction rate.
    $\Psi>0$ whenever r=$\|g_{\mathrm{bd}}\|_2/\|g_{\mathrm{cl}}\|_2$ exceeds a threshold $r^\star(\kappa,\lambda_{\mathrm{bd}})$, which holds at the start of Phase~3 when $\mathcal{L}_{\mathrm{bd}}$ is large and $\mathcal{L}_{\mathrm{cl}}$ is at its minimum.
\end{theorem}

With MI alignment ($\kappa > 0$), Part~(a) guarantees joint descent of the clean and backdoor loss. Part~(b) ensures that enforcing MI alignment makes $\Phi(\theta_1^\star)$ decrease at rate $\Psi$ when 
the gradient ratio $r=\|g_{\mathrm{bd}}\|_2/\|g_{\mathrm{cl}}\|_2$ 
exceeds the threshold $r^\star(\kappa,\lambda_{\mathrm{bd}})$, which is 
at most $\max(1,\,1/\!\sqrt{\lambda_{\mathrm{bd}}})$ for all 
$\kappa\in[0,1]$.
This condition holds naturally at the start of Phase~3: 
$\mathcal{L}_{\mathrm{bd}}$ is large because the trigger--target 
association is not yet implanted, while $\mathcal{L}_{\mathrm{cl}}$ 
is at its minimum after clean pre-training, giving $r \gg r^\star$.

Combining Theorem~\ref{thm:mi-gap} with the anti-forgetting guarantee on $\mathcal{L}_{\mathrm{cl}}$ (see Assumption~\ref{ass:af} in Appendix), we can derive an upper bound on the backdoor loss after $t$-1 benign downstream finetuning updates.

\begin{theorem}[Free-Rider Persistence of BadGCL]\label{cor:freerider}
    Under the condition of Theorem~\ref{thm:mi-gap} and the Lipschitz smoothness assumption of the GNN model, we further assume bounded representation perturbation, gradient perturbation induced by backdoor poisoning and trigger-node features (Assumptions 2 and 4 in Appendix~\ref {sec:theory-assumptions}). These assumptions hold for most practically deployed GNN models and backdoor attack scenarios. 
    We use $\delta_{\mathrm{cl}}$ to denote the cumulative clean-loss degradation tolerated by the GCL system. The backdoor loss after $t{-}1$ benign downstream updates satisfies
    \begin{equation}\label{eq:freerider}
    \begin{split}
        &\mathcal{L}_{\mathrm{bd}}(\theta_t)
        \;\le\;
        \mathcal{L}_{\mathrm{cl}}(\theta_1^\star)
        +\delta_{\mathrm{cl}}
        \;+\;
        \Phi(\theta_1^\star) +\;L\textstyle\sum_{i}\|\Delta\theta_i\|_2^2\\
        \;+\;
        &\bigl(C_{\mathrm{feat}}
            +C_{\mathrm{str}}\sqrt{m\rho}
            +C_{\mathrm{prop}}\sqrt{\epsilon_{\mathrm{prop}}}
            +C_{\mathrm{label}}
            +2L D_t\bigr)
            \textstyle\sum_{i}\|\Delta\theta_i\|_2\\
    \end{split}
    \end{equation}
where $\delta_{\mathrm{cl}} \geq 0$ is the clean task loss
degradation induced but constrained by the GCL anti-forgetting mechanism. $\epsilon_{\mathrm{prop}}=\mathcal{L}_{\mathrm{prop}}(G_\Delta^\star)$ is the propagation loss of the optimized trigger, $m$ and $\rho$ are the size and edge density of the trigger subgraph, $C_{\mathrm{feat}}=C_g C_x B_X$ is the trigger-feature contribution, $C_{\mathrm{str}}=C_g C_A$ is the adjacency contribution, $C_{\mathrm{prop}}$ captures propagation instability, and $C_{\mathrm{label}}=\|\tilde{g}_{\mathrm{cl}}(\theta_{\mathrm{cl}})-g_{\mathrm{cl}}(\theta_{\mathrm{cl}})\|_2$ is the label mismatch between the anchor-set gradient under the target label and the full clean gradient (see Lemma~\ref{lem:mismatch-ref} for exact definitions). $\Delta\theta_i = \theta_{i+1} - \theta_i$ denotes the parameter update at the i-th downstream GCL step. $C_g$, $C_x$, $C_A$ and $B_x$ are defined as the constants bounding the representation perturbation, gradient perturbation and trigger-node features (see Assumption 2 and 4 in Appendix~\ref {sec:theory-assumptions})
\end{theorem}

Theorem~\ref{cor:freerider} formalizes the \emph{free-rider persistence} of BadGCL. Specifically, 
$C_{\mathrm{prop}}\sqrt{\epsilon_{\mathrm{prop}}}$ bounds the drift from propagation instability ($\mathcal{L}_{\mathrm{prop}}$), $C_{\mathrm{str}}\sqrt{m\rho}$ from the trigger's topological footprint ($\mathcal{R}_{\mathrm{str}}$), $C_{\mathrm{feat}}$ from the trigger-feature magnitude, and $C_{\mathrm{label}}$ from the label mismatch.
The MI objective $\mathcal{L}_{\mathrm{MI}}$ in BadGCL serves a dual role: it minimizes the \emph{initial} gap $\Phi(\theta_1^\star)$ before downstream benign updates (Theorem~\ref{thm:mi-gap}) and indirectly drives $C_{\mathrm{label}}$ small by aligning the backdoor gradient with the clean-task signal (Lemma~\ref{lem:mi-sep}).

As shown, every term on the right-hand side of Eq.~\eqref{eq:freerider} is either a quantity the GCL system already preserves ($\mathcal{L}_{\mathrm{cl}}(\theta_1^\star)+\delta_{\mathrm{cl}}$ is the cumulative clean-loss budget that anti-forgetting permits) or a quantity the attacker minimizes in Phase~2 ($\Phi(\theta_1^\star)$ via $\mathcal{L}_{\mathrm{MI}}$, $\epsilon_{\mathrm{prop}}$ via $\mathcal{L}_{\mathrm{prop}}$, $m\rho$ via $\mathcal{R}_{\mathrm{str}}$, $C_{\mathrm{feat}}$ via the trigger-feature bound and $C_{\mathrm{label}}$ via the MI alignment). 
$\sum_i\|\Delta\theta_i\|_2$ stays marginally small, which is implicitly enforced 
by GCL anti-forgetting mechanisms: replay-based methods limit updates 
through historical sample losses, regularization-based methods penalize 
parameter changes in important directions, and parameter-isolation 
methods freeze shared parameters entirely.
Therefore, when all three objectives of BadGCL are well-optimized, the backdoor loss is bounded by $\mathcal{L}_{\mathrm{cl}}(\theta_1^\star)+\delta_{\mathrm{cl}}$. 
Later updates that preserve clean first-task knowledge (keeping $\delta_{\mathrm{cl}}$ small) therefore \emph{inadvertently also preserve the backdoor}, producing the persistence effect validated in Section~\ref{sec:experiments}. 


\section{Experiments}
\label{sec:experiments}
\begin{table*}[t]
\centering
\footnotesize
\setlength{\abovecaptionskip}{3pt}
\setlength{\belowcaptionskip}{0pt}
\setlength{\tabcolsep}{1.5pt}
\renewcommand{\arraystretch}{1.10}
\caption{Overall final performance on the \textsc{CoraFull} and \textsc{Reddit} datasets under the task-IL protocol across three victim GCL systems (all values in \%). \emph{w/o} poisoning denotes the benign reference for each victim system. The benign reference is shown for comparison only and excluded from best/runner-up ranking; best and runner-up results are highlighted in \textbf{bold} and \underline{underline}, respectively. $\uparrow$ indicates higher is better, while $\downarrow$ indicates lower is better.}
\label{tab:main_taskIL}
\begin{tabular}{l||cccc||cccc||cccc||cccc||cccc||cccc}
\toprule
Victim & \multicolumn{8}{c||}{\textsc{PDGNNs~\cite{PDGNNs}}} & \multicolumn{8}{c||}{\textsc{CaT-CGL}~\cite{CaT-CGL}} & \multicolumn{8}{c}{\textsc{SSRM}~\cite{SSRM}}\\
\cmidrule(lr){1-1} \cmidrule(lr){2-9} \cmidrule(lr){10-17} \cmidrule(lr){18-25}
\multirow{2.5}{*}{Method}
& \multicolumn{4}{c||}{\textsc{CoraFull}}
& \multicolumn{4}{c||}{\textsc{Reddit}}
& \multicolumn{4}{c||}{\textsc{CoraFull}}
& \multicolumn{4}{c||}{\textsc{Reddit}} & \multicolumn{4}{c||}{\textsc{CoraFull}}
& \multicolumn{4}{c}{\textsc{Reddit}}\\
\cmidrule(lr){2-5}\cmidrule(lr){6-9} \cmidrule(lr){10-13} \cmidrule(lr){14-17} \cmidrule(lr){18-21} \cmidrule(lr){22-25} 
& {AA}$\uparrow$ & {AF}$\downarrow$ & {ASR}$\uparrow$ & {BF}$\downarrow$
& {AA}$\uparrow$ & {AF}$\downarrow$ & {ASR}$\uparrow$ & {BF}$\downarrow$ 
& {AA}$\uparrow$ & {AF}$\downarrow$ & {ASR}$\uparrow$ & {BF}$\downarrow$ 
& {AA}$\uparrow$ & {AF}$\downarrow$ & {ASR}$\uparrow$ & {BF}$\downarrow$
& {AA}$\uparrow$ & {AF}$\downarrow$ & {ASR}$\uparrow$ & {BF}$\downarrow$ 
& {AA}$\uparrow$ & {AF}$\downarrow$ & {ASR}$\uparrow$ & {BF}$\downarrow$ \\
\midrule

\textit{w/o} poisoning
& 92.76 & 0.78 & {--} & {--}
& 97.92 & 1.07 & {--} & {--}
& 91.68 & 0.64 & {--} & {--}
& 96.91 & 0.20 & {--} & {--}
& 87.75 & 3.00 & {--} & {--}
& 95.34 & 2.61 & {--} & {--} \\[-1.5ex]

\multicolumn{25}{c}{\tikz \draw[densely dashed] (0,0) -- (1\linewidth,0);} \\

Sub--BD~\cite{sub_backdoor}
& 72.46 & 19.21 & 55.38 & 44.62
& 80.84 & 18.51 & 58.12 & 41.88
& 70.91 & 20.44 & 52.06 & 47.94
& 78.35 & 17.66 & 56.47 & 43.53
& 68.74 & 23.18 & 48.91 & 51.09
& 76.28 & 21.36 & 54.72 & 45.28 \\

GTA~\cite{GTA}
& 72.62 & 17.91 & 49.82 & 50.18
& 79.79 & 12.59 & 60.32 & 39.68
& 71.36 & 19.12 & 53.21 & 46.79
& 78.92 & 14.05 & 60.51 & 39.49
& 69.20 & 21.57 & 46.83 & 53.17
& 75.91 & 18.44 & 57.35 & 42.65 \\

Motif--BD~\cite{motif-backdoor}
& 81.58 & 10.23 & 51.42 & 48.58
& 78.64 & 18.70 & 55.45 & 44.55
& 80.42 & 11.05 & 50.38 & 49.62
& 77.83 & 19.24 & 55.02 & 44.98
& 78.36 & 14.77 & 49.56 & 50.44
& 74.62 & 22.15 & 53.64 & 46.36 \\

UGBA~\cite{UGBA}
& 84.80 & 12.82 & 54.48 & 45.52
& 74.85 & 23.39 & 60.12 & 39.88
& 82.76 & 13.38 & 55.03 & 44.97
& 73.94 & 24.72 & 59.16 & 40.84
& 80.15 & 16.06 & 52.78 & 47.22
& 72.48 & 26.10 & 57.08 & 42.92 \\

NFTA~\cite{NFTA}
& 84.34 & 13.22 & 51.32 & 48.68
& 76.73 & 21.00 & 64.23 & 35.77
& 83.18 & 13.64 & 52.67 & 47.33
& 75.82 & 22.48 & 63.74 & 36.26
& 82.45 & 14.58 & 50.33 & 49.67
& 73.66 & 24.03 & 60.29 & 39.71 \\

GCBA~\cite{GCBA}
& \underline{86.39} & 11.46 & 55.12 & 44.88
& 86.54 & 11.30 & 67.85 & 32.15
& \underline{85.07} & 12.14 & 56.38 & 43.62
& 84.92 & 12.41 & 68.73 & 31.27
& 79.54 & 16.91 & 54.07 & 45.93
& 82.15 & 15.37 & 65.42 & 34.58 \\

CGBA~\cite{CGBA}
& 74.34 & 23.51 & 53.15 & 46.85
& 86.62 & 10.19 & 60.88 & 39.12
& 73.55 & 24.82 & 54.86 & 45.14
& 81.38 & 14.54 & 61.39 & 38.61
& 71.84 & 27.66 & 50.74 & 49.26
& 83.02 & 14.72 & 58.94 & 41.06 \\

SPEAR~\cite{SPEAR}
& 84.78 & 12.79 & 52.32 & 47.68
& 84.40 & 13.30 & 62.45 & 37.55
& 83.71 & 13.26 & 53.88 & 46.12
& 83.21 & 14.52 & 63.22 & 36.78
& 80.82 & 16.12 & 50.91 & 49.09
& 81.04 & 16.38 & 61.18 & 38.82 \\

\midrule

LTB~\cite{latent}
& 81.64 & 6.88 & 83.15 & 16.85
& 89.86 & 7.25 & 82.12 & 17.88
& 81.96 & 7.34 & 84.58 & 15.42
& 88.42 & 8.16 & 83.67 & 16.33
& 79.84 & 9.26 & 77.91 & 22.09
& \underline{86.17} & 8.78 & 80.24 & 19.76 \\

P-Trojan~\cite{P-Trojan}
& 85.27 & \underline{5.43} & \underline{89.36} & \underline{10.64}
& \underline{92.04} & \underline{5.71} & \underline{88.93} & \underline{11.07}
& 84.34 & \underline{6.16} & \underline{91.24} & \underline{8.76}
& \underline{90.61} & \underline{5.94} & \underline{90.37} & \underline{9.63}
& \underline{83.52} & \underline{8.38} & \underline{84.65} & \underline{15.35}
& 85.43 & \underline{7.51} & \underline{87.62} & \underline{12.38} \\

\midrule

\sysname{}
& \textbf{92.54} & \textbf{0.92} & \textbf{99.46} & \textbf{0.54}
& \textbf{97.76} & \textbf{1.02} & \textbf{98.17} & \textbf{1.83} & \textbf{91.65} & \textbf{0.43} & \textbf{97.12} & \textbf{2.38} & \textbf{94.52} & \textbf{0.31} & \textbf{99.86} & \textbf{0.14} & \textbf{86.58} & \textbf{6.94} & \textbf{97.91} & \textbf{2.09} & \textbf{94.77} & \textbf{2.37} & \textbf{97.25} &  \textbf{2.14} \\

\bottomrule
\end{tabular}
\vspace{-3.5mm}
\end{table*}

To verify Observation~\ref{obs:persistence_in_gcl} and evaluate \sysname{}, we organize our experiments around the following research questions:
\begin{itemize}[leftmargin=*]
    \item \textbf{\textit{RQ1)}} Can \sysname{} achieve stealthy and persistent backdoor attacks across GCL protocols, datasets, and victim systems?
    \item \textbf{\textit{RQ2)}} How does \sysname{} preserve clean utility and backdoor effectiveness throughout continual learning?
    \item \textbf{\textit{RQ3)}} Is the persistence of \sysname{} robust to task-order changes and benign task revisiting?
    \item \textbf{\textit{RQ4)}} How resilient is \sysname{} against graph-backdoor defenses?
    \item \textbf{\textit{RQ5)}} How do the key objectives in \sysname{} contribute to persistence, robustness, and clean-utility preservation?
\end{itemize}

\subsection{Experimental Setup}

\noindent \textbf{Victim algorithms.}
We evaluate \sysname{} on four representative GCL victim systems: replay-based PDGNNs~\cite{PDGNNs} and CaT-CGL~\cite{CaT-CGL}, regularization-based SSRM~\cite{SSRM}, and parameter-isolation-based TPP~\cite{TPP}.
Details of victim systems are provided in Appendix~\ref{app:victim_systems}.


\noindent \textbf{Datasets.} We use four standard GCL benchmark datasets, \textsc{CoraFull}~\cite{CoraFull}, \textsc{OGB-Arxiv}~\cite{Arxiv}, \textsc{Reddit}~\cite{Reddit}, and \textsc{OGB-Products}~\cite{Arxiv}, following common task splits where each task contains two classes. This yields 35 tasks (70 classes) on \textsc{CoraFull}, 20 tasks (40 classes) on \textsc{OGB-Arxiv}, 20 tasks (40 classes) on \textsc{Reddit}, and 23 tasks (47 classes) on \textsc{OGB-Products}. On each of the datasets, we assume the attacker performs backdoor poisoning only over the first task.

\noindent \textbf{Baselines.}
Existing backdoor attacks are not specifically designed for GCL, so we adapt attacks from two related domains as baselines.
We consider: \textit{(i) representative graph backdoor attacks} including Sub-BD~\cite{sub_backdoor}, GTA~\cite{GTA}, Motif-BD~\cite{motif-backdoor}, UGBA~\cite{UGBA}, NFTA~\cite{NFTA}, GCBA~\cite{GCBA}, CGBA~\cite{CGBA}, and SPEAR~\cite{SPEAR}, which are originally designed for static graph learning; and \textit{(ii) continual-learning backdoor attacks for non-graph modalities} including Latent Task Backdoor (LTB)~\cite{latent} and P-Trojan~\cite{P-Trojan}, which we adapt to graph learning by replacing their original backbones with GNN encoders while preserving their core backdoor mechanisms.
All baselines are evaluated under the same threat model as \sysname{}: first-task intervention with poisoning budget $\epsilon$ and no adversarial access to later tasks.

\noindent\textbf{Evaluation metrics.}
We evaluate clean GCL performance using the accuracy matrix and backdoor performance.
We report final metrics after the last task: \emph{Average Accuracy} (AA) for clean utility, \emph{Average Forgetting} (AF) for clean-task forgetting, \emph{Attack Success Rate} (ASR) for backdoor effectiveness, and \emph{Backdoor Forgetting} (BF) for ASR degradation across continual updates.
For AA and ASR, higher is better ($\uparrow$); for AF and BF, lower is better ($\downarrow$).
All values are reported in \%, averaged over five independent runs.
More details are provided in Appendix~\ref{sec:appendix_exp_metrics}.

\begin{table*}[t]
\centering
\footnotesize
\setlength{\abovecaptionskip}{3pt}
\setlength{\belowcaptionskip}{0pt}
\setlength{\tabcolsep}{1.5pt}
\renewcommand{\arraystretch}{1.10}
\caption{Overall final performance on the \textsc{CoraFull}, \textsc{Reddit}, and \textsc{OGB-Arxiv} datasets under the class-IL protocol across two victim GCL systems (all values in \%). 
}
\label{tab:main_classIL}
\begin{tabular}{l||cccc||cccc||cccc||cccc||cccc||cccc}
\toprule
Victim & \multicolumn{12}{c||}{\textsc{PDGNNs}~\cite{PDGNNs}} & \multicolumn{12}{c}{\textsc{CaT-CGL}~\cite{CaT-CGL}}\\
\cmidrule(lr){1-1} \cmidrule(lr){2-13} \cmidrule(lr){14-25}
\multirow{2.3}{*}{Method}
& \multicolumn{4}{c||}{\textsc{CoraFull}}
& \multicolumn{4}{c||}{\textsc{Reddit}}
& \multicolumn{4}{c||}{\textsc{OGB-Arxiv}} & \multicolumn{4}{c||}{\textsc{CoraFull}}
& \multicolumn{4}{c||}{\textsc{Reddit}}
& \multicolumn{4}{c}{\textsc{OGB-Arxiv}} \\
\cmidrule(lr){2-5}\cmidrule(lr){6-9}\cmidrule(lr){10-13} \cmidrule(lr){14-17}\cmidrule(lr){18-21}\cmidrule(lr){22-25}
& {AA}$\uparrow$ & {AF}$\downarrow$ & {ASR}$\uparrow$ & {BF}$\downarrow$
& {AA}$\uparrow$ & {AF}$\downarrow$ & {ASR}$\uparrow$ & {BF}$\downarrow$
& {AA}$\uparrow$ & {AF}$\downarrow$ & {ASR}$\uparrow$ & {BF}$\downarrow$ & {AA}$\uparrow$ & {AF}$\downarrow$ & {ASR}$\uparrow$ & {BF}$\downarrow$
& {AA}$\uparrow$ & {AF}$\downarrow$ & {ASR}$\uparrow$ & {BF}$\downarrow$
& {AA}$\uparrow$ & {AF}$\downarrow$ & {ASR}$\uparrow$ & {BF}$\downarrow$ \\
\midrule

\textit{w/o} poisoning
& 78.12 & 5.04 & {--} & {--}
& 95.12 & 2.45 & {--} & {--}
& 52.23 & 13.25 & {--} & {--}
& 67.34 & 5.70 & {--} & {--}
& 94.54 & 1.96 & {--} & {--}
& 45.96 & 12.71 & {--} & {--} \\[-1.7ex]

\multicolumn{25}{c}{\tikz \draw[densely dashed] (0,0) -- (1\linewidth,0);} \\

Sub--BD~\cite{sub_backdoor}
& 13.83 & 68.42 & 3.12 & 96.34
& 23.41 & 67.96 & 5.45 & 94.55
& 2.14 & 78.41 & 4.63 & 95.37
& 12.46 & 59.88 & 4.21 & 95.79
& 24.08 & 66.73 & 6.02 & 93.98
& 3.72 & 76.18 & 6.41 & 93.59 \\

GTA~\cite{GTA}
& 16.44 & 65.15 & 3.05 & 96.95
& 26.97 & 68.81 & 5.42 & 94.58
& 4.05 & 75.12 & 5.21 & 94.79
& 15.92 & 56.44 & 5.16 & 94.84
& 22.15 & 69.04 & 5.31 & 94.69
& 5.06 & 73.82 & 5.88 & 94.12 \\

Motif--BD~\cite{motif-backdoor}
& 15.12 & 67.12 & 2.98 & 97.02
& 27.14 & 65.72 & 5.35 & 94.65
& 1.81 & 79.10 & 5.01 & 94.99
& 13.58 & 58.91 & 4.04 & 95.96
& 28.76 & 63.42 & 5.72 & 94.28
& 2.68 & 77.54 & 6.07 & 93.93 \\

UGBA~\cite{UGBA}
& 15.85 & 66.12 & 3.05 & 96.95
& 27.87 & 68.08 & 5.42 & 94.58
& 3.15 & 77.44 & 4.86 & 95.14
& 17.24 & 55.83 & 4.83 & 95.17
& 25.61 & 68.95 & 6.18 & 93.82
& 3.49 & 75.06 & 5.42 & 94.58 \\

NFTA~\cite{NFTA}
& 17.43 & 64.11 & 3.12 & 96.88
& 28.88 & 67.20 & 5.38 & 94.62
& 5.82 & 74.15 & 5.58 & 94.42
& 18.06 & 54.72 & 5.47 & 94.53
& 29.34 & 64.83 & 6.04 & 93.96
& 6.15 & 71.68 & 6.56 & 93.44 \\

GCBA~\cite{GCBA}
& 18.81 & 62.45 & 3.02 & 96.98
& 30.81 & 64.99 & 5.45 & 94.55
& 6.43 & 72.82 & 5.48 & 94.52
& 19.35 & 53.61 & 5.09 & 94.91
& 31.72 & 61.45 & 6.33 & 93.67
& 6.74 & 70.89 & 5.93 & 94.07 \\

CGBA~\cite{CGBA}
& 16.87 & 65.49 & 3.08 & 96.92
& 26.97 & 68.99 & 5.95 & 94.05
& 4.45 & 76.15 & 5.35 & 94.65
& 16.73 & 57.28 & 5.63 & 94.37
& 27.09 & 67.16 & 6.85 & 93.15
& 4.38 & 74.21 & 6.22 & 93.78 \\

SPEAR~\cite{SPEAR}
& 17.15 & 63.88 & 3.05 & 96.95
& 28.25 & 70.42 & 5.88 & 94.12
& 4.82 & 75.88 & 5.78 & 94.22
& 17.88 & 55.47 & 4.91 & 95.09
& 30.14 & 66.05 & 6.74 & 93.26
& 5.27 & 72.96 & 6.71 & 93.29 \\

\midrule

LTB~\cite{latent}
& 39.26 & 52.11 & 2.98 & 97.02
& 43.03 & 52.44 & 5.45 & 94.55
& 19.79 & 41.83 & 5.48 & 94.52
& 36.18 & 43.79 & 5.28 & 94.72
& 41.69 & 50.86 & 6.19 & 93.81
& 18.44 & 39.77 & 6.02 & 93.98 \\

P-Trojan~\cite{P-Trojan}
& \underline{58.73} & \underline{30.28} & \underline{67.86} & \underline{32.14}
& \underline{62.64} & \underline{34.72} & \underline{59.35} & \underline{40.65}
& \underline{29.38} & \underline{33.42} & \underline{58.16} & \underline{38.73}
& \underline{49.52} & \underline{33.31} & \underline{61.74} & \underline{35.12}
& \underline{50.87} & \underline{37.18} & \underline{63.06} & \underline{36.72}
& \underline{26.91} & \underline{33.64} & \underline{59.88} & \underline{38.24} \\

\midrule

\sysname{}
& \textbf{77.93} & \textbf{4.92} & \textbf{98.74} & \textbf{1.08}
& \textbf{95.12} & \textbf{2.39} & \textbf{95.48} & \textbf{3.93}
& \textbf{52.08} & \textbf{13.67} & \textbf{95.65} & \textbf{4.11} & \textbf{66.36} & \textbf{5.42} & \textbf{95.94} & \textbf{3.03} & \textbf{92.85} & \textbf{2.17} & \textbf{99.86} & \textbf{0.09} & \textbf{45.83} & \textbf{12.79} & \textbf{99.28} & \textbf{0.72}\\

\bottomrule
\end{tabular}
\vspace{-2mm}
\end{table*}


\subsection{Overall Performance (RQ1)}
\label{sec:overall_performance}

Tables~\ref{tab:main_taskIL} and~\ref{tab:main_classIL} summarize the main results, while additional results are deferred to Appendix~\ref{sec:appendix_exp} due to space limitations.
Across both protocols and victim systems, \sysname{} consistently achieves the strongest backdoor persistence while maintaining clean GCL utility close to the benign reference.
Under task-IL, \sysname{} obtains near-saturated ASR on replay-based victims and remains effective on the regularization-based SSRM, while keeping AA close to the corresponding \emph{w/o} poisoning model.
Under class-IL, where the model must discriminate over the shared accumulated label space, \sysname{} still maintains high final ASR and low BF on both PDGNNs and CaT-CGL.
These results support our central motivation: a persistent GCL backdoor cannot rely only on an initial trigger-label association; it must couple the poisoned update with the clean-task signal preserved by GCL and stabilize the trigger effect through graph message passing.

The comparison with baselines further confirms this point.
Static graph backdoor attacks usually achieve limited ASR and large BF, especially under class-IL, because their triggers are not designed to survive continual updates.
Continual-learning backdoor baselines such as LTB and P-Trojan improve over static attacks, but they are adapted from non-graph modalities and do not explicitly model graph propagation.
As a result, they still lag behind \sysname{} in both persistence and clean utility.
The consistent gains across replay-based and regularization-based victims indicate that \sysname{} is not exploiting a single GCL implementation, but rather the broader preservation behavior shared by GCL systems.

\begin{figure}
    \centering
    
    \includegraphics[width=0.9\linewidth]{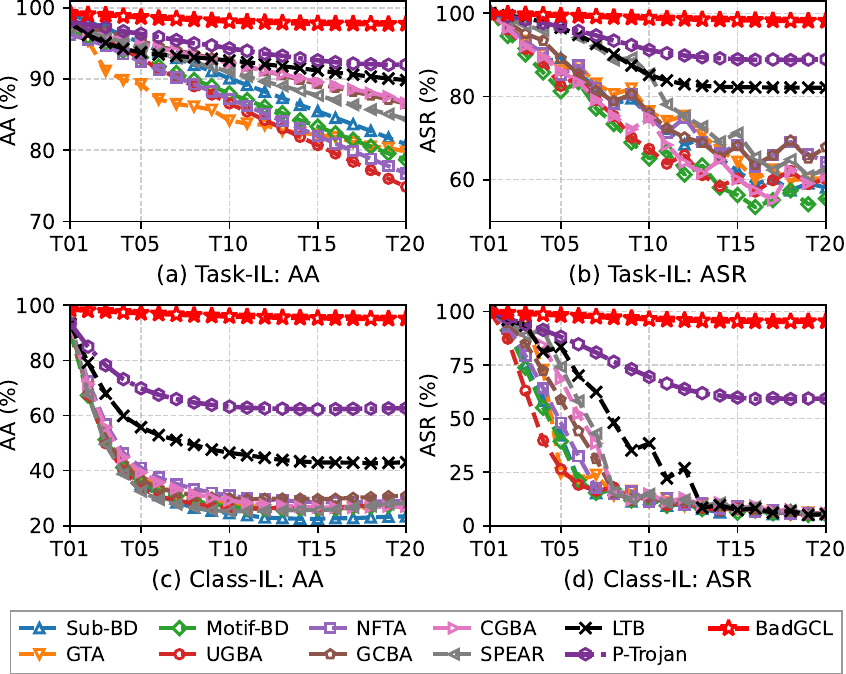}
    \setlength{\abovecaptionskip}{3pt}
    \setlength{\belowcaptionskip}{-5pt}
    \caption{Performance evolution ($\mathrm{AA}$ and $\mathrm{ASR}$ across tasks) on \textsc{Reddit} dataset under the task-IL and class-IL protocols.}
    \label{fig:task_change}
    \vspace{-2mm}
\end{figure}

\subsection{Performance Dynamics (RQ2)}

We further examine the dynamics of clean utility and backdoor effectiveness over tasks with PDGNNs, as shown in Figure~\ref{fig:task_change}.
Most baselines follow a common trajectory: the backdoor is effective in early tasks, gradually decays as continual learning proceeds, and eventually becomes ineffective, with a sharper collapse under the more challenging class-IL protocol.
LTB delays this degradation to some extent, and P-Trojan provides a stronger continual-learning backdoor baseline; however, both remain clearly behind \sysname{}, especially in preserving ASR while maintaining clean AA.
In contrast, \sysname{} keeps high ASR throughout the continual trajectory while its clean utility stays close to the benign learning trend.
This supports our design motivation: $\mathcal{L}_{\mathrm{MI}}$ couples the poisoned gradient with the clean-task signal preserved by GCL, while $\mathcal{L}_{\mathrm{prop}}$ stabilizes the trigger-induced effect through graph message passing.
The stable co-evolution of AA and ASR further corroborates the persistence paradox in Observation~\ref{obs:persistence_in_gcl}: mechanisms intended to preserve historical knowledge can also preserve a well-optimized backdoor.
We further analyze task-wise clean-utility retention through accuracy matrices in Appendix~\ref{sec:appendix_exp_accuracy}.


\begin{table}[t]
\centering
\footnotesize
\setlength{\abovecaptionskip}{3pt}
\setlength{\belowcaptionskip}{0pt}
\setlength{\tabcolsep}{4.5pt}
\renewcommand{\arraystretch}{1.08}
\caption{Task-sequence robustness on \textsc{Reddit} with PDGNNs.}
\label{tab:task_sequence_change}
\begin{tabular}{lcc||cc}
\toprule
\multirow{2.5}{*}{\makecell[c]{Task\\Sequence}}
& \multicolumn{2}{c||}{Task-IL}
& \multicolumn{2}{c}{Class-IL} \\
\cmidrule(lr){2-3}\cmidrule(lr){4-5}
& AA (\%) $\uparrow$ & ASR (\%) $\uparrow$
& AA (\%) $\uparrow$ & ASR (\%) $\uparrow$ \\
\midrule
Original
& 97.76 & 98.17
& 95.12 & 95.48 \\
Reverse
& 97.72 & 98.36
& 94.96 & 95.15 \\
Random (fixed first)
& 97.79{\scriptsize$\pm$0.04} & 98.14{\scriptsize$\pm$0.14}
& 94.98{\scriptsize$\pm$0.08} & 95.56{\scriptsize$\pm$0.20} \\
Full Random
& 97.73{\scriptsize$\pm$0.12} & 97.54{\scriptsize$\pm$0.80}
& 94.74{\scriptsize$\pm$0.23} & 92.98{\scriptsize$\pm$0.53} \\
\bottomrule
\end{tabular}
\vspace{-2mm}
\end{table}

\begin{figure}
    \centering
    \setlength{\abovecaptionskip}{3pt}
    \setlength{\belowcaptionskip}{-5pt}
    \includegraphics[width=0.9\linewidth]{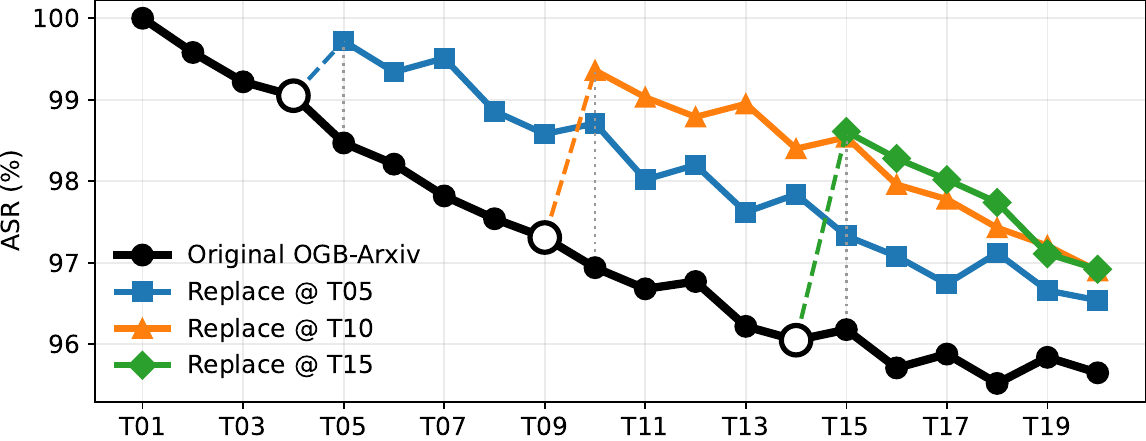}
    \caption{ASR trajectories under task replacement at different continual stages for class-IL \textsc{OGB-Arxiv} dataset. The dashed lines denote the new trajectories after task replacement.}
    \label{fig:asr_change}
    \vspace{-3mm}
\end{figure}

\subsection{Robustness to Task Variations (RQ3)}

\noindent \textbf{Robustness to task-sequence changes.}
We evaluate task-order sensitivity on \textsc{Reddit} with PDGNNs, as shown in Table~\ref{tab:task_sequence_change}.
We compare the original order with the reverse order, ten post-first-task random orders that keep the poisoned first task unchanged, and ten full-random orders where the whole task sequence is shuffled and the new first task is poisoned.
Across both task-IL and class-IL settings, \sysname{} maintains stable AA and ASR under reverse and post-first-task random orders, with only minor fluctuations from the original sequence.
Under full-random orders, where the poisoned task itself also changes, \sysname{} still preserves high ASR with limited clean-utility degradation.
These results indicate that the persistence of \sysname{} is not tied to a specific hand-picked task order or poisoned task, but generalizes across substantially different continual learning sequences.

\noindent \textbf{Effect of task revisiting on backdoor persistence.}
To further examine whether the implanted backdoor behaves like preserved historical knowledge, we conduct a task-replacement experiment on \textsc{OGB-Arxiv} with PDGNNs.
For the 20-task stream, we consider four settings: the original sequence and three variants where the 5th, 10th, or 15th task is replaced by the poisoned first task, and compare ASR trajectories in Figure~\ref{fig:asr_change}.
Whenever replacement occurs, ASR rises to varying extents, indicating that benign revisiting of the poisoned task can re-consolidate the implanted backdoor.
This supports our motivation: by coupling the poisoned gradient with the poisoned-task clean optimization signal and stabilizing the trigger effect through message passing, \sysname{} causes the backdoor to be reinforced by later GCL updates that revisit or preserve historical knowledge.
Such benign revisits therefore induce a replay-like reinforcement effect, further strengthening persistent backdoors without any additional adversarial intervention.

\begin{table}[t]
\footnotesize
\centering
\setlength{\abovecaptionskip}{3pt}
\setlength{\belowcaptionskip}{-5pt}
\setlength{\tabcolsep}{4.5pt}
\renewcommand{\arraystretch}{1.15}
\caption{Defense and ablation results on \textsc{CoraFull} with PDGNNs. (a) evaluates defenses against \sysname{}; (b) tests structure regularization under defenses; (c) tests propagation ablation without defenses. Subscripts indicate the absolute change relative to the corresponding attacked model.}
\label{tab:defense_corafull}
\begin{tabular}{l cc cc}
\toprule
\quad\quad\quad\quad Model
& \multicolumn{2}{c}{Task-IL} & \multicolumn{2}{c}{Class-IL} \\
\specialrule{0pt}{-1pt}{0pt}
\cmidrule(lr){2-3} \cmidrule(lr){4-5}
\specialrule{0pt}{-0.5pt}{0pt}
(\atk Model / \defn Defense) & AA (\%) $\uparrow$ & ASR (\%) $\uparrow$ & AA (\%) $\uparrow$ & ASR (\%) $\uparrow$\\
\midrule

\multicolumn{5}{c}{\textbf{(a) Countermeasures against \sysname{}}} \\
\midrule

\atk \sysname{}
& 92.54 & 99.46 & 77.93 & 98.74 \\

\\[-4.7ex]
\multicolumn{5}{c}{\tikz \draw[densely dashed] (0,0) -- (0.95\linewidth,0);} \\
[-0.5ex]
\defn \quad AGNNCert~\cite{Agnncert}
& 92.38\drop{0.16} & 99.28\drop{0.18} & 77.81\drop{0.12} & 98.46\drop{0.28} \\
\defn \quad AACL~\cite{AACL}
& 89.96\drop{2.58} & 96.82\drop{2.64} & 74.62\drop{3.31} & 96.34\drop{2.40} \\
\defn \quad OD~\cite{OD}
& 91.87\drop{0.67} & 96.50\drop{2.96} & 76.58\drop{1.35} & 95.57\drop{3.17} \\
\defn \quad RIGBD~\cite{RIGBD}
& 90.92\drop{1.62} & 94.92\drop{4.54} & 75.24\drop{2.69} & 94.15\drop{4.59} \\
\defn \quad $\text{A}^2$GBD~\cite{A2GBD}
& 84.31\drop{8.23} & 91.76\drop{7.70} & 68.86\drop{9.07} & 89.42\drop{9.32} \\

\midrule
\midrule

\multicolumn{5}{c}{\textbf{(b) Structure regularization under countermeasures}} \\
\midrule

\atk \sysname{} \textit{w/o} $\mathcal{R}_{\mathrm{str}}$
& 92.61 & 99.31 & 78.12 & 98.52 \\

\\[-4.7ex]
\multicolumn{5}{c}{\tikz \draw[densely dashed] (0,0) -- (0.95\linewidth,0);} \\
[-0.5ex]
\defn \quad AGNNCert~\cite{Agnncert}
& 92.40\drop{0.21} & 99.12\drop{0.19} & 77.86\drop{0.26} & 98.34\drop{0.18} \\
\defn \quad AACL~\cite{AACL}
& 89.42\drop{3.19} & 95.38\drop{3.93} & 72.31\drop{5.81} & 93.46\drop{5.06} \\
\defn \quad OD~\cite{OD}
& 91.18\drop{1.43} & 93.86\drop{5.45} & 75.42\drop{2.70} & 90.64\drop{7.88} \\
\defn \quad RIGBD~\cite{RIGBD}
& 89.66\drop{2.95} & 88.74\drop{10.57} & 73.56\drop{4.56} & 69.82\drop{28.70} \\
\defn \quad $\text{A}^2$GBD~\cite{A2GBD}
& 81.78\drop{10.83} & 86.91\drop{12.40} & 65.22\drop{12.90} & 83.36\drop{15.16} \\

\midrule
\midrule

\multicolumn{5}{c}{\textbf{(c) Propagation ablation}} \\
\midrule

\sysname{} \textit{w/o} $\mathcal{L}_{\mathrm{prop}}$
& 92.16 & 95.84 & 76.81 & 92.46 \\

\bottomrule
\end{tabular}
\vspace{-4mm}
\end{table}




\subsection{Potential Countermeasures (RQ4)}

We evaluate \sysname{} against five representative graph-backdoor defenses.
As shown in Table~\ref{tab:defense_corafull}(a), AGNNCert~\cite{Agnncert} has little effect because inference-time aggregation changes cannot remove the backdoor consolidated during early training.
AACL~\cite{AACL}, OD~\cite{OD}, and RIGBD~\cite{RIGBD} reduce ASR to varying degrees, but \sysname{} remains effective due to its structurally controlled and clean-compatible trigger.
$\text{A}^2$GBD~\cite{A2GBD} gives the strongest mitigation, but also incurs the largest AA drop, weakening clean GCL utility.
Overall, existing defenses either have limited impact or require non-negligible utility degradation for stronger mitigation.
Detailed defense descriptions are provided in Appendix~\ref{sec:appendix_defense_details}.

\begin{figure}
    \centering
    \setlength{\abovecaptionskip}{3pt}
    \setlength{\belowcaptionskip}{-5pt}
    \includegraphics[width=0.95\linewidth]{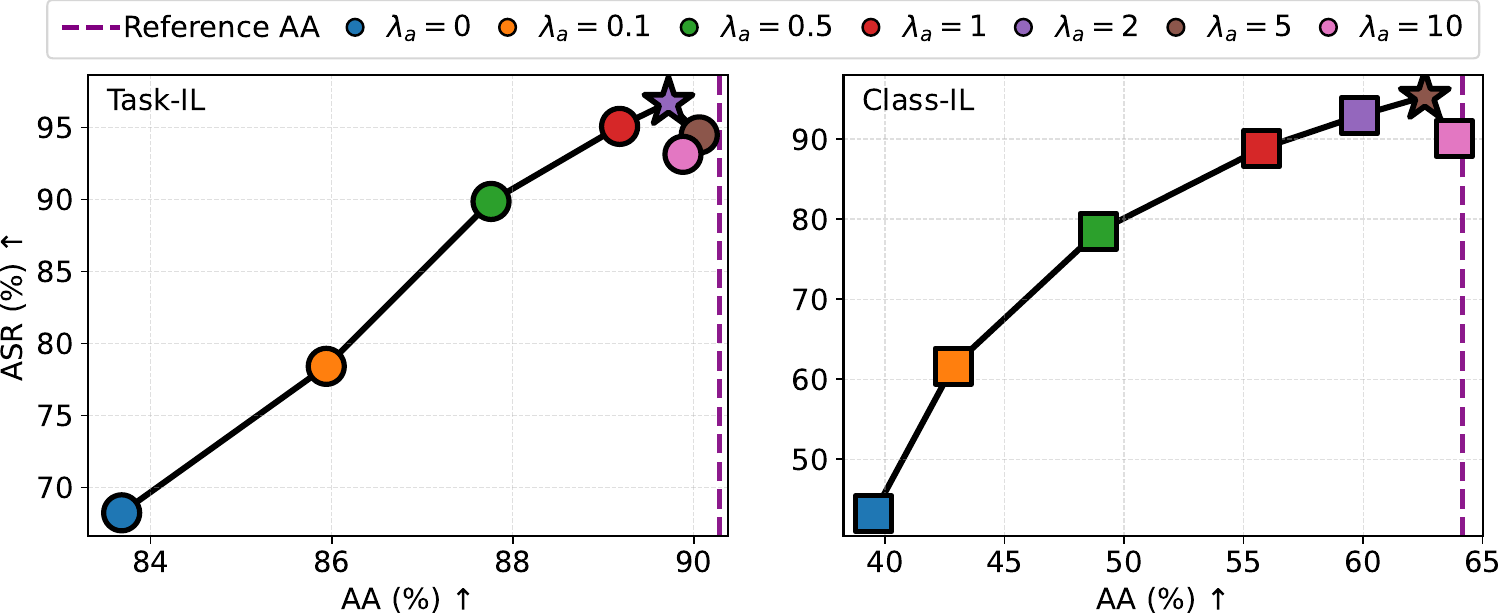}
    \caption{Ablation of the gradient-compatibility weight $\lambda_{\mathrm{MI}}$ on \textsc{OGB-Products} with PDGNNs. Points closer to the upper-right corner indicate better trade-offs between clean utility and persistent backdoor effectiveness. The star marks the selected weight, and the dashed vertical line indicates the reference AA of benign PDGNNs.}
    \label{fig:lambda_tradeoff}
    \vspace{-2mm}
\end{figure}

\subsection{Ablation Study (RQ5)}

\noindent \textbf{Structure regularization.}
Table~\ref{tab:defense_corafull}(b) evaluates the effect of removing the trigger structure regularizer $\mathcal{R}_{\mathrm{str}}$.
Without this regularizer, \sysname{} remains effective without defenses, but becomes substantially more vulnerable to structure-oriented countermeasures.
The gap is especially clear under RIGBD and $\text{A}^2$GBD, where ASR drops much more sharply than the full model.
This confirms that $\mathcal{R}_{\mathrm{str}}$ controls the trigger's structural footprint and improves its robustness against pruning or edge manipulation.

\noindent \textbf{Propagation objective.}
Table~\ref{tab:defense_corafull}(c) studies the effect of removing the propagation objective $\mathcal{L}_{\mathrm{prop}}$.
The AA drop is mild, while ASR decreases clearly under both GCL protocols, especially under class-IL.
This indicates that $\mathcal{L}_{\mathrm{prop}}$ is not mainly responsible for clean-task fitting; instead, it stabilizes the trigger-induced representation shift through graph message passing, which is critical for preserving the backdoor after later benign updates.

\noindent \textbf{MI weight.}
To quantify the balance between immediate backdoor strength and long-term persistence, we vary $\lambda_{\mathrm{MI}} \in \{0, 0.1, 0.5, 1, 2, 5, \\10\}$ on \textsc{OGB-Products} with PDGNNs, as shown in Figure~\ref{fig:lambda_tradeoff}.
When $\lambda_{\mathrm{MI}}=0$, the attack loses its CL-aware compatibility constraint and becomes closer to static graph backdoor optimization, leading to weak persistence.
Moderately increasing $\lambda_{\mathrm{MI}}$ improves ASR by coupling the poisoned gradient with the clean-task signal preserved by GCL, while overly large values weaken direct backdoor fitting.
This verifies that persistent GCL backdoors require balancing immediate attack strength with compatibility to future continual updates; accordingly, we set $\lambda_{\mathrm{MI}}=2$ for task-IL and $\lambda_{\mathrm{MI}}=5$ for class-IL.

\section{Conclusion}
\label{sec:conclusion}
In this work, we identify a critical security vulnerability in graph continual learning (GCL): mechanisms for preserving historical knowledge can also preserve early injected backdoor associations.
We propose \sysname{}, a model-agnostic framework that constructs gradient-compatible structural triggers to align backdoor optimization with GCL knowledge-preservation dynamics while stabilizing trigger-induced effects through graph message passing.
Our theory shows that gradient compatibility and propagation stability jointly support long-term backdoor persistence under continual updates.
Across task-IL and class-IL protocols, multiple datasets, and diverse victim systems, \sysname{} achieves persistent backdoor effectiveness with limited clean-utility degradation and remains robust to task-level variations, hyperparameter changes, and representative graph-backdoor defenses.
These findings expose a security blind spot in GCL and call for defenses that distinguish benign historical knowledge from malicious structural associations consolidated during continual learning.

\section*{Acknowledgment}
This work was supported by the National Science Foundation (NSF) \# 2515265.


\bibliographystyle{ACM-Reference-Format}
\bibliography{kdd26}

@String{Springer = "Springer-Verlag" }

@inproceedings{sub_backdoor,
  title={Backdoor attacks to graph neural networks},
  author={Zhang, Zaixi and Jia, Jinyuan and Wang, Binghui and Gong, Neil Zhenqiang},
  booktitle={Proceedings of the 26th ACM symposium on access control models and technologies},
  pages={15--26},
  year={2021}
}

@inproceedings{GTA,
  title={Graph backdoor},
  author={Xi, Zhaohan and Pang, Ren and Ji, Shouling and Wang, Ting},
  booktitle={30th USENIX security symposium (USENIX Security 21)},
  pages={1523--1540},
  year={2021}
}

@article{motif-backdoor,
  title={Motif-backdoor: Rethinking the backdoor attack on graph neural networks via motifs},
  author={Zheng, Haibin and Xiong, Haiyang and Chen, Jinyin and Ma, Haonan and Huang, Guohan},
  journal={IEEE Transactions on Computational Social Systems},
  volume={11},
  number={2},
  pages={2479--2493},
  year={2023},
  publisher={IEEE}
}

@inproceedings{UGBA,
  title={Unnoticeable backdoor attacks on graph neural networks},
  author={Dai, Enyan and Lin, Minhua and Zhang, Xiang and Wang, Suhang},
  booktitle={Proceedings of the ACM Web Conference 2023},
  pages={2263--2273},
  year={2023}
}

@article{NFTA,
  title={Feature-Based Graph Backdoor Attack in the Node Classification Task},
  author={Chen, Yang and Ye, Zhonglin and Zhao, Haixing and Wang, Ying},
  journal={International Journal of Intelligent Systems},
  volume={2023},
  number={1},
  pages={5418398},
  year={2023},
  publisher={Wiley Online Library}
}

@inproceedings{GCBA,
  title={Graph contrastive backdoor attacks},
  author={Zhang, Hangfan and Chen, Jinghui and Lin, Lu and Jia, Jinyuan and Wu, Dinghao},
  booktitle={International Conference on Machine Learning},
  pages={40888--40910},
  year={2023},
  organization={PMLR}
}

@article{CGBA,
  title={A clean-label graph backdoor attack method in node classification task},
  author={Xing, Xiaogang and Xu, Ming and Bai, Yujing and Yang, Dongdong},
  journal={Knowledge-Based Systems},
  volume={304},
  pages={112433},
  year={2024},
  publisher={Elsevier}
}

@inproceedings{SPEAR,
  title={SPEAR: A Structure-Preserving Manipulation Method for Graph Backdoor Attacks},
  author={Ding, Yuanhao and Liu, Yang and Ji, Yugang and Wen, Weigao and He, Qing and Ao, Xiang},
  booktitle={Proceedings of the ACM on Web Conference 2025},
  pages={1237--1247},
  year={2025}
}

@inproceedings{latent,
  title={Persistent backdoor attacks in continual learning},
  author={Guo, Zhen and Kumar, Abhinav and Tourani, Reza},
  booktitle={34th USENIX Security Symposium (USENIX Security 25)},
  pages={6379--6397},
  year={2025}
}

@inproceedings{PDGNNs,
  title={Topology-aware embedding memory for continual learning on expanding networks},
  author={Zhang, Xikun and Song, Dongjin and Chen, Yixin and Tao, Dacheng},
  booktitle={Proceedings of the 30th ACM SIGKDD Conference on Knowledge Discovery and Data Mining},
  pages={4326--4337},
  year={2024}
}

@article{CoraFull,
  title={Automating the construction of internet portals with machine learning},
  author={McCallum, Andrew Kachites and Nigam, Kamal and Rennie, Jason and Seymore, Kristie},
  journal={Information Retrieval},
  volume={3},
  number={2},
  pages={127--163},
  year={2000},
  publisher={Springer}
}

@article{Arxiv,
  title={Open graph benchmark: Datasets for machine learning on graphs},
  author={Hu, Weihua and Fey, Matthias and Zitnik, Marinka and Dong, Yuxiao and Ren, Hongyu and Liu, Bowen and Catasta, Michele and Leskovec, Jure},
  journal={Advances in neural information processing systems},
  volume={33},
  pages={22118--22133},
  year={2020}
}

@article{Reddit,
  title={Inductive representation learning on large graphs},
  author={Hamilton, Will and Ying, Zhitao and Leskovec, Jure},
  journal={Advances in neural information processing systems},
  volume={30},
  year={2017}
}

@article{survey,
  title={Continual learning on graphs: Challenges, solutions, and opportunities},
  author={Zhang, Xikun and Song, Dongjin and Tao, Dacheng},
  journal={arXiv preprint arXiv:2402.11565},
  year={2024}
}

@article{CBACL,
  title={Red alarm: Controllable backdoor attack in continual learning},
  author={Gao, Rui and Liu, Weiwei},
  journal={Neural Networks},
  pages={107479},
  year={2025},
  publisher={Elsevier}
}

@article{nguyen2024backdoor,
  title={Backdoor attack in prompt-based continual learning},
  author={Nguyen, Trang and Tran, Anh and Ho, Nhat},
  journal={arXiv e-prints},
  pages={arXiv--2406},
  year={2024}
}

@article{graph_survey,
  title={Graph neural backdoor: Fundamentals, methodologies, applications, and future directions},
  author={Yang, Xiao and Li, Gaolei and Li, Jianhua},
  journal={arXiv preprint arXiv:2406.10573},
  year={2024}
}

@article{PruneG,
  title={Adversarial examples on graph data: Deep insights into attack and defense},
  author={Wu, Huijun and Wang, Chen and Tyshetskiy, Yuriy and Docherty, Andrew and Lu, Kai and Zhu, Liming},
  journal={arXiv preprint arXiv:1903.01610},
  year={2019}
}

@inproceedings{Agnncert,
  title={Agnncert: Defending graph neural networks against arbitrary perturbations with deterministic certification},
  author={Li, Jiate and Wang, Binghui},
  booktitle={USENIX Security},
  volume={1},
  number={2},
  pages={3},
  year={2025}
}

@article{AACL,
  title={Adversary aware continual learning},
  author={Umer, Muhammad and Polikar, Robi},
  journal={IEEE Access},
  year={2024},
  publisher={IEEE}
}

@article{kipf2016semi,
  title={Semi-supervised classification with graph convolutional networks},
  author={Kipf, TN},
  journal={arXiv preprint arXiv:1609.02907},
  year={2016}
}

@inproceedings{xi2021graph,
  title={Graph backdoor},
  author={Xi, Zhaohan and Pang, Ren and Ji, Shouling and Wang, Ting},
  booktitle={30th USENIX security symposium (USENIX Security 21)},
  pages={1523--1540},
  year={2021}
}

@inproceedings{zhang2024rethinking,
  title={Rethinking graph backdoor attacks: A distribution-preserving perspective},
  author={Zhang, Zhiwei and Lin, Minhua and Dai, Enyan and Wang, Suhang},
  booktitle={Proceedings of the 30th ACM SIGKDD Conference on Knowledge Discovery and Data Mining},
  pages={4386--4397},
  year={2024}
}

@inproceedings{xu2021explainability,
  title={Explainability-based backdoor attacks against graph neural networks},
  author={Xu, Jing and Xue, Minhui and Picek, Stjepan},
  booktitle={Proceedings of the 3rd ACM workshop on wireless security and machine learning},
  pages={31--36},
  year={2021}
}

@article{li2022backdoor,
  title={Backdoor learning: A survey},
  author={Li, Yiming and Jiang, Yong and Li, Zhifeng and Xia, Shu-Tao},
  journal={IEEE transactions on neural networks and learning systems},
  volume={35},
  number={1},
  pages={5--22},
  year={2022},
  publisher={IEEE}
}

@article{gao2020backdoor,
  title={Backdoor attacks and countermeasures on deep learning: A comprehensive review},
  author={Gao, Yansong and Doan, Bao Gia and Zhang, Zhi and Ma, Siqi and Zhang, Jiliang and Fu, Anmin and Nepal, Surya and Kim, Hyoungshick},
  journal={arXiv preprint arXiv:2007.10760},
  year={2020}
}

@article{wang2024comprehensive,
  title={A comprehensive survey of continual learning: Theory, method and application},
  author={Wang, Liyuan and Zhang, Xingxing and Su, Hang and Zhu, Jun},
  journal={IEEE transactions on pattern analysis and machine intelligence},
  volume={46},
  number={8},
  pages={5362--5383},
  year={2024},
  publisher={IEEE}
}

@article{TPP,
  title={Replay-and-forget-free graph class-incremental learning: A task profiling and prompting approach},
  author={Niu, Chaoxi and Pang, Guansong and Chen, Ling and Liu, Bing},
  journal={Advances in Neural Information Processing Systems},
  volume={37},
  pages={87978--88002},
  year={2024}
}

@inproceedings{SSRM,
  title={Towards robust graph incremental learning on evolving graphs},
  author={Su, Junwei and Zou, Difan and Zhang, Zijun and Wu, Chuan},
  booktitle={International Conference on Machine Learning},
  pages={32728--32748},
  year={2023},
  organization={PMLR}
}

@inproceedings{P-Trojan,
  title={Persistent backdoor attacks under continual fine-tuning of llms},
  author={Cui, Jing and Han, Yufei and Jiao, Jianbin and Zhang, Junge},
  booktitle={Proceedings of the AAAI Conference on Artificial Intelligence},
  volume={40},
  number={36},
  pages={30422--30430},
  year={2026}
}

@inproceedings{CaT-CGL,
  title={Cat: Balanced continual graph learning with graph condensation},
  author={Liu, Yilun and Qiu, Ruihong and Huang, Zi},
  booktitle={2023 IEEE International Conference on Data Mining (ICDM)},
  pages={1157--1162},
  year={2023},
  organization={IEEE}
}

@inproceedings{OD,
  title={Rethinking graph backdoor attacks: A distribution-preserving perspective},
  author={Zhang, Zhiwei and Lin, Minhua and Dai, Enyan and Wang, Suhang},
  booktitle={Proceedings of the 30th ACM SIGKDD conference on knowledge discovery and data mining},
  pages={4386--4397},
  year={2024}
}

@article{RIGBD,
  title={Robustness-inspired defense against backdoor attacks on graph neural networks},
  author={Zhang, Zhiwei and Lin, Minhua and Xu, Junjie and Wu, Zongyu and Dai, Enyan and Wang, Suhang},
  journal={arXiv preprint arXiv:2406.09836},
  year={2024}
}

@inproceedings{A2GBD,
  title={A2GBD: Attack-Agnostic Graph Backdoor Defense},
  author={Du, Chenxu and Liu, Yang and Yu, Xingtong and Xu, Zhuoer and Liu, Yang and Li, Tianrui},
  booktitle={Proceedings of the ACM Web Conference 2026},
  pages={1229--1239},
  year={2026}
}

@inproceedings{DShield,
  title={DShield: Defending against Backdoor Attacks on Graph Neural Networks via Discrepancy Learning.},
  author={Yu, Hao and Ma, Chuan and Wan, Xinhang and Wang, Jun and Xiang, Tao and Shen, Meng and Liu, Xinwang},
  booktitle={NDSS},
  year={2025}
}

@article{Oord2018RepresentationLW,
  title={Representation Learning with Contrastive Predictive Coding},
  author={A{\"a}ron van den Oord and Yazhe Li and Oriol Vinyals},
  journal={ArXiv},
  year={2018},
  volume={abs/1807.03748},
  url={https://api.semanticscholar.org/CorpusID:49670925}
}

@article{goodfellow2013empirical,
  title={An empirical investigation of catastrophic forgetting in gradient-based neural networks},
  author={Goodfellow, Ian J and Mirza, Mehdi and Xiao, Da and Courville, Aaron and Bengio, Yoshua},
  journal={arXiv preprint arXiv:1312.6211},
  year={2013}
}

@inproceedings{poole2019variational,
  title     = {On Variational Bounds of Mutual Information},
  author    = {Poole, Ben and Ozair, Sherjil and van den Oord, A{\"a}ron 
               and Alemi, Alexander A. and Tucker, George},
  booktitle = {Proceedings of the 36th International Conference on 
               Machine Learning (ICML)},
  volume    = {97},
  pages     = {5171--5180},
  year      = {2019},
  publisher = {PMLR}
}

@inproceedings{li2023contrastive,
  title={Contrastive representation Learning based on multiple Node-centered Subgraphs},
  author={Li, Dong and Wang, Wenjun and Shao, Minglai and Zhao, Chen},
  booktitle={Proceedings of the 32nd ACM International Conference on Information and Knowledge Management},
  pages={1338--1347},
  year={2023}
}

@inproceedings{tian2025mldgg,
  title={Mldgg: Meta-learning for domain generalization on graphs},
  author={Tian, Qin and Zhao, Chen and Shao, Minglai and Wang, Wenjun and Lin, Yujie and Li, Dong},
  booktitle={Proceedings of the 31st ACM SIGKDD Conference on Knowledge Discovery and Data Mining V. 1},
  pages={1361--1372},
  year={2025}
}

@inproceedings{ma2026llm,
  title={LLM-enhanced energy contrastive learning for out-of-distribution detection in text-attributed graphs},
  author={Ma, Xiaoxu and Li, Dong and Shao, Minglai and Wu, Xintao and Zhao, Chen},
  booktitle={Proceedings of the AAAI Conference on Artificial Intelligence},
  volume={40},
  number={29},
  pages={24308--24316},
  year={2026}
}

@inproceedings{li2026marlin,
  title={MARLIN: Multi-Agent Reinforcement Learning for Incremental DAG Discovery},
  author={Li, Dong and Chen, Zhengzhang and Zhao, Xujiang and Yu, Linlin and Chen, Zhong and He, Yi and Chen, Haifeng and Zhao, Chen},
  booktitle={Proceedings of the AAAI Conference on Artificial Intelligence},
  volume={40},
  number={27},
  pages={22869--22877},
  year={2026}
}

@article{li2026solverllm,
  title={SolverLLM: Leveraging test-time scaling for optimization problem via LLM-guided search},
  author={Li, Dong and Zhao, Xujiang and Yu, Linlin and Liu, Yanchi and Cheng, Wei and Chen, Zhengzhang and Chen, Zhong and Chen, Feng and Zhao, Chen and Chen, Haifeng},
  journal={Advances in Neural Information Processing Systems},
  volume={38},
  pages={100028--100058},
  year={2026}
}

@article{li2026hierflow,
  title={HierFlow: Hierarchical Coupled Dual-Space Search for Automatic Agentic Workflow Generation},
  author={Li, Dong and Liu, Yanchi and Zhao, Xujiang and Cheng, Wei and Chen, Zhengzhang and Wu, Xintao and Chen, Zhong and Chen, Haifeng and Zhao, Chen},
  journal={arXiv preprint arXiv:2607.21609},
  year={2026}
}

@inproceedings{tian2026class,
  title={Class-Domain Incremental Learning on Graphs via Disentangled Knowledge Distillation},
  author={Tian, Qin and Zhao, Chen and Wu, Xintao and Li, Dong and Shao, Minglai and Zhao, Xujiang and Wang, Wenjun},
  booktitle={Proceedings of the ACM Web Conference 2026},
  pages={452--462},
  year={2026}
}

@inproceedings{zhao2025uncertainty,
  title={Uncertainty propagation on llm agent},
  author={Zhao, Qiwei and Li, Dong and Liu, Yanchi and Cheng, Wei and Sun, Yiyou and Oishi, Mika and Osaki, Takao and Matsuda, Katsushi and Yao, Huaxiu and Zhao, Chen and others},
  booktitle={Proceedings of the 63rd Annual Meeting of the Association for Computational Linguistics (Volume 1: Long Papers)},
  pages={6064--6073},
  year={2025}
}

@article{shao2024supervised,
  title={Supervised algorithmic fairness in distribution shifts: A survey},
  author={Shao, Minglai and Li, Dong and Zhao, Chen and Wu, Xintao and Lin, Yujie and Tian, Qin},
  journal={arXiv preprint arXiv:2402.01327},
  year={2024}
}

\appendix
\setcounter{theorem}{0}

\section{Detailed Theoretical Analysis of \sysname{}}
\label{sec:theory}
 
This appendix provides the full theoretical development supporting the main results in Section~\ref{sec:feasibility}. We state all definitions, assumptions, and supporting lemmas, and give complete proofs of Theorems~\ref{thm:mi-gap} and ~\ref{cor:freerider}.
 
\subsection{Definitions}\label{sec:theory-defs}
 
\begin{definition}[Clean and Backdoor Losses]\label{def:losses}
    The \emph{clean first-task loss} and the \emph{backdoor loss} at parameters $\theta$ are
    \begin{align}
        \mathcal{L}_{\mathrm{cl}}(\theta)
            &\;=\;\mathbb{E}_{(x,y)\sim\mathcal{D}_1^{\mathrm{tr}}}
                \bigl[\ell\bigl(f_\theta(x),\,y\bigr)\bigr],
            \label{eq:Lcl}\\[2pt]
        \mathcal{L}_{\mathrm{bd}}(\theta)
            &\;=\;\mathbb{E}_{(x,y)\sim\mathcal{A}}
                \bigl[\ell\bigl(f_\theta\bigl(\mathcal{I}(x,G_\Delta)\bigr),\,y^\star\bigr)\bigr],
            \label{eq:Lbd}
    \end{align}
    where $\ell$ is the cross-entropy loss. The corresponding gradients are denoted $g_{\mathrm{cl}}(\theta)=\nabla_\theta\mathcal{L}_{\mathrm{cl}}(\theta)$ and $g_{\mathrm{bd}}(\theta)=\nabla_\theta\mathcal{L}_{\mathrm{bd}}(\theta)$.
\end{definition}
 
\begin{definition}[Target-Task Loss Gap]\label{def:gap}
    The \emph{target-task loss gap} at parameters $\theta$ is $\Phi(\theta)=\mathcal{L}_{\mathrm{bd}}(\theta)-\mathcal{L}_{\mathrm{cl}}(\theta)$.
    A small $\Phi(\theta)$ means the backdoor loss tracks the clean first-task loss: if GCL preserves $\mathcal{L}_{\mathrm{cl}}(\theta_t)$, a small gap implies $\mathcal{L}_{\mathrm{bd}}(\theta_t)$ is also preserved.
\end{definition}
 
\begin{definition}[Gradient Mismatch]\label{def:mismatch}
    The \emph{gradient mismatch} at parameters $\theta$ is $\delta g(\theta)=g_{\mathrm{bd}}(\theta)-g_{\mathrm{cl}}(\theta)=\nabla_\theta\Phi(\theta)$.
    When $\|\delta g(\theta)\|_2$ is small along the parameter trajectory, the gap $\Phi$ changes slowly under downstream updates.
\end{definition}
 
\begin{definition}[Trigger-Induced Propagation Shift]\label{def:prop-shift}
    For an anchor node $v\in\mathcal{A}$, let $h_v^{(k)}$ and $h_{\Delta,v}^{(k)}$ denote the node representation after the $k$-th message-passing layer before and after trigger injection. The \emph{trigger-induced shift} at depth $k$ is $r_\Delta^{(k)}(v)=h_{\Delta,v}^{(k)}-h_v^{(k)}$.
\end{definition}
 
\begin{definition}[Cosine Alignment]\label{def:cosine}
    For nonzero vectors $a,b\in\mathbb{R}^d$, the cosine similarity is $s(a,b)=\langle a,b\rangle/(\|a\|_2\|b\|_2)$. We write $\kappa=s(g_{\mathrm{bd}},g_{\mathrm{cl}})$ for the gradient alignment at $\theta_{\mathrm{cl}}$.
\end{definition}
 
\subsection{Assumptions}\label{sec:theory-assumptions}
 
We use a minimal set of local regularity conditions that hold along the GCL trajectory.
 
\begin{assumption}[Smooth Losses]\label{ass:smooth}
    Both $\mathcal{L}_{\mathrm{cl}}$ and $\mathcal{L}_{\mathrm{bd}}$ are differentiable in $\theta$, with $\mathcal{L}_{\mathrm{bd}}$ being $L_{\mathrm{bd}}$-smooth and $\mathcal{L}_{\mathrm{cl}}$ being $L_{\mathrm{cl}}$-smooth:
    $\|\nabla_\theta\mathcal{L}_\bullet(\theta)-\nabla_\theta\mathcal{L}_\bullet(\theta')\|_2\le L_\bullet\|\theta-\theta'\|_2$ for $\bullet\in\{\mathrm{cl},\mathrm{bd}\}$, for all $\theta,\theta'$ in a convex neighborhood of the trajectory $\{\theta_1^\star,\ldots,\theta_T\}$. We write $L=\max(L_{\mathrm{bd}},L_{\mathrm{cl}})$.
\end{assumption}
 
\begin{assumption}[Lipschitz Message Passing]\label{ass:lipschitz-mp}
    The GNN propagation and backpropagation maps are Lipschitz around the anchor-local computation graph. Concretely, there exist constants $C_x,C_A>0$ such that the representation perturbation caused by inserting a trigger with node features $X_\Delta$ and adjacency $A_\Delta$ satisfies, for every anchor node $v$ and every propagation depth $k\le K$,
    \begin{equation}\label{eq:lip-mp}
        \|r_\Delta^{(k)}(v)\|_2 \;\le\; C_x\|X_\Delta\|_F + C_A\|A_\Delta\|_F.
    \end{equation}
    Moreover, perturbation bounds on propagated representations transfer to gradient perturbation bounds up to a constant factor $C_g>0$:
    \begin{equation}\label{eq:lip-grad}
        \bigl\|g_{\mathrm{bd}}(\theta)-\tilde{g}_{\mathrm{cl}}(\theta)\bigr\|_2 \;\le\; C_g\;\max_{v\in\mathcal{A}}\|r_\Delta^{(K)}(v)\|_2,
    \end{equation}
    where $\tilde{g}_{\mathrm{cl}}(\theta)=\mathbb{E}_{(x,y)\sim\mathcal{A}}[\nabla_\theta\ell(f_\theta(x),y^\star)]$ is the gradient of the loss on \emph{clean} anchor inputs evaluated against the target label $y^\star$.
\end{assumption}
 
\begin{assumption}[Propagation Stability]\label{ass:prop}
    The optimized trigger satisfies $\mathcal{L}_{\mathrm{prop}}(G_\Delta^\star)\le\epsilon_{\mathrm{prop}}$, i.e., the trigger-induced representation shift remains directionally and magnitudinally consistent across message-passing depths $k=1,\ldots,K$.
\end{assumption}
 
\begin{assumption}[Structural Sparsity]\label{ass:sparse}
    After the density regularizer $\mathcal{R}_{\mathrm{str}}$, the trigger $G_\Delta^\star$ has $m$ nodes and edge density $\rho$. Under degree normalization, $\|A_\Delta\|_F\le\sqrt{m\rho}$. The trigger-node features are bounded: $\|X_\Delta\|_F\le B_X$.
\end{assumption}
 
\begin{assumption}[Effective Anti-Forgetting]\label{ass:af}
    At each downstream step $i\ge 1$, the GCL mechanism preserves a first-task \emph{protected subspace} $\mathcal{U}_i$ containing $g_{\mathrm{cl}}(\theta_i)$: for all $u\in\mathcal{U}_i$,
    \begin{equation}\label{eq:anti-forget}
        \langle u,\;\Delta\theta_i\rangle \;\le\; \tau_i\,\|u\|_2,
    \end{equation}
    where $\tau_i\ge 0$ is the effective anti-forgetting tolerance at step $i$. Summing over the trajectory yields a cumulative clean-loss preservation guarantee:
    \begin{equation}\label{eq:cl-preserve}
        \mathcal{L}_{\mathrm{cl}}(\theta_t) \;\le\; \mathcal{L}_{\mathrm{cl}}(\theta_1^\star) \;+\;\delta_{\mathrm{cl}},
    \end{equation}
    where $\delta_{\mathrm{cl}}=\sum_{i=1}^{t-1}\tau_i\|g_{\mathrm{cl}}(\theta_i)\|_2+\tfrac{L}{2}\sum_{i=1}^{t-1}\|\Delta\theta_i\|_2^2$ is the cumulative degradation the GCL system permits.
\end{assumption}
 
\begin{remark}
    Assumption~\ref{ass:af} is a unified abstraction of anti-forgetting. Replay-based methods (e.g.\ PDGNNs) enforce Eq.~\eqref{eq:anti-forget} by constraining updates that would increase the replay loss; regularization-based methods (e.g.\ SSRM) penalize parameter changes in important directions; prompt-based methods (e.g.\ TPP) freeze the backbone and learn task-specific prompts, making $\Delta\theta_i$ structurally constrained. In each case, $\tau_i$ captures how tightly the mechanism protects historical knowledge.
\end{remark}
 
\subsection{Supporting Lemmas}\label{sec:theory-lemmas}
 
\begin{lemma}[Gradient Mismatch Bound at the Reference Point]\label{lem:mismatch-ref}
    Under Assumptions~\ref{ass:lipschitz-mp}--\ref{ass:sparse}, the gradient mismatch at the clean reference $\theta_{\mathrm{cl}}$ decomposes as
    \begin{equation}\label{eq:mismatch-decomp}
        \delta g(\theta_{\mathrm{cl}})
        \;=\;
        \underbrace{\bigl[g_{\mathrm{bd}}(\theta_{\mathrm{cl}})-\tilde{g}_{\mathrm{cl}}(\theta_{\mathrm{cl}})\bigr]}_{\text{structural mismatch}}
        \;+\;
        \underbrace{\bigl[\tilde{g}_{\mathrm{cl}}(\theta_{\mathrm{cl}})-g_{\mathrm{cl}}(\theta_{\mathrm{cl}})\bigr]}_{\text{label mismatch}},
    \end{equation}
    where $\tilde{g}_{\mathrm{cl}}(\theta)=\mathbb{E}_{(x,y)\sim\mathcal{A}}[\nabla_\theta\ell(f_\theta(x),y^\star)]$ as in Assumption~\ref{ass:lipschitz-mp}. The structural mismatch satisfies
    \begin{equation}\label{eq:struct-bound}
        \bigl\|g_{\mathrm{bd}}(\theta_{\mathrm{cl}})-\tilde{g}_{\mathrm{cl}}(\theta_{\mathrm{cl}})\bigr\|_2
        \;\le\;
        C_{\mathrm{feat}} + C_{\mathrm{str}}\sqrt{m\rho} + C_{\mathrm{prop}}\sqrt{\epsilon_{\mathrm{prop}}},
    \end{equation}
    where $C_{\mathrm{feat}}=C_g C_x B_X$ captures the trigger-feature contribution, $C_{\mathrm{str}}=C_g C_A$ captures the adjacency contribution, and \\$C_{\mathrm{prop}} = C_g \left( B\sqrt{N} + \sqrt{N/\gamma} \right)$ captures the additional gradient perturbation from propagation instability. Defining the label mismatch as $C_{\mathrm{label}}=\bigl\|\tilde{g}_{\mathrm{cl}}(\theta_{\mathrm{cl}})-g_{\mathrm{cl}}(\theta_{\mathrm{cl}})\bigr\|_2$, the total gradient mismatch at the reference point satisfies
    \begin{equation}\label{eq:total-mismatch}
        \|\delta g(\theta_{\mathrm{cl}})\|_2
        \;\le\;
        C_{\mathrm{feat}} + C_{\mathrm{str}}\sqrt{m\rho} + C_{\mathrm{prop}}\sqrt{\epsilon_{\mathrm{prop}}} + C_{\mathrm{label}}.
    \end{equation}
\end{lemma}
 
\begin{proof}
    The structural mismatch is the gradient difference caused solely by replacing the clean input $x$ with the triggered input $\mathcal{I}(x,G_\Delta)$, holding the target label $y^\star$ fixed. By Assumption~\ref{ass:lipschitz-mp}, Eq.~\eqref{eq:lip-grad}, this difference is bounded by $C_g\max_{v\in\mathcal{A}}\|r_\Delta^{(K)}(v)\|_2$. We decompose $r_\Delta^{(K)}(v) = r_\Delta^{(1)}(v) + [r_\Delta^{(K)}(v) - r_\Delta^{(1)}(v)]$ via the triangle inequality. For the first-layer shift, Eq.~\eqref{eq:lip-mp} and Assumption~\ref{ass:sparse} give $\|r_\Delta^{(1)}(v)\|_2 \le C_x B_X + C_A\sqrt{m\rho}$, yielding the $C_{\mathrm{feat}} + C_{\mathrm{str}}\sqrt{m\rho}$ terms after multiplication by $C_g$. For the propagation deviation, Assumption~\ref{ass:prop} guarantees that the layer-wise shifts $r_\Delta^{(k)}(v)$ remain close to $r_\Delta^{(1)}(v)$. By Lemma~\ref{lem:prop-stable}, the deviation $\|r_\Delta^{(K)}(v)-r_\Delta^{(1)}(v)\|_2$ is bounded by $\mathcal{O}(\sqrt{\epsilon_{\mathrm{prop}}})$. Through the Lipschitz backpropagation transfer, this induces an additional gradient perturbation of at most $C_{\mathrm{prop}}\sqrt{\epsilon_{\mathrm{prop}}}$. The label mismatch $C_{\mathrm{label}}$ follows from the triangle inequality applied to Eq.~\eqref{eq:mismatch-decomp}.
\end{proof}
 
\begin{lemma}[Propagation Stability Bound]\label{lem:prop-stable}
    Let $N=|\mathcal{A}|(K-1)$. Under Assumption~\ref{ass:prop}, if $\mathcal{L}_{\mathrm{prop}}(G_\Delta)\le\epsilon_{\mathrm{prop}}$ and $\|r_\Delta^{(1)}(v)\|_2\le B$ for every anchor node, then for any $v\in\mathcal{A}$ and $k\in\{2,\ldots,K\}$,
    \begin{equation}\label{eq:prop-stable}
        \bigl\|r_\Delta^{(k)}(v)-r_\Delta^{(1)}(v)\bigr\|_2
        \;\le\; B\sqrt{N\epsilon_{\mathrm{prop}}} + \sqrt{{N\epsilon_{\mathrm{prop}}}/{\gamma}} + o\!\bigl(\sqrt{\epsilon_{\mathrm{prop}}}\bigr),
    \end{equation}
    where $\gamma>0$ is the magnitude-consistency weight in $\mathcal{L}_{\mathrm{prop}}$.
\end{lemma}
 
\begin{proof}
    The propagation loss averages nonnegative direction and magnitude terms over all anchors and depths. Since every summand is nonneg.\ and the average is at most $\epsilon_{\mathrm{prop}}$, each individual direction discrepancy satisfies $d_{\mathrm{dir}}(r_\Delta^{(k)}(v),r_\Delta^{(1)}(v))\le N\epsilon_{\mathrm{prop}}$, and each magnitude discrepancy satisfies $\bigl|\|r_\Delta^{(k)}(v)\|_2-\|r_\Delta^{(1)}(v)\|_2\bigr|\le\sqrt{N\epsilon_{\mathrm{prop}}/\gamma}$.
    Writing $r_\Delta^{(k)}(v)=\|r_\Delta^{(k)}(v)\|_2\cdot\hat{r}_\Delta^{(k)}(v)$ and applying the triangle inequality,
    \begin{align*}
        \|r_\Delta^{(k)}(v)-r_\Delta^{(1)}(v)\|_2
        &\le \|r_\Delta^{(1)}(v)\|_2\cdot\|\hat{r}_\Delta^{(k)}(v)-\hat{r}_\Delta^{(1)}(v)\|_2 \\
        &+ \bigl|\|r_\Delta^{(k)}(v)\|_2-\|r_\Delta^{(1)}(v)\|_2\bigr|\\
        &\le B\sqrt{N\epsilon_{\mathrm{prop}}} + \sqrt{N\epsilon_{\mathrm{prop}}/\gamma} + o(\sqrt{\epsilon_{\mathrm{prop}}}),
    \end{align*}
    where the first term uses $d_{\mathrm{dir}}(a,b)=\|\hat{a}-\hat{b}\|_2^2\le N\epsilon_{\mathrm{prop}}$ and $\|r_\Delta^{(1)}(v)\|_2\le B$.
\end{proof}
 
\begin{lemma}[Gradient Mismatch Along the Trajectory]\label{lem:mismatch-traj}
    Under Assumption~\ref{ass:smooth}, for any $\theta_i$ on the GCL trajectory,
    \begin{equation}\label{eq:mismatch-traj}
        \|\delta g(\theta_i)\|_2 \;\le\; \|\delta g(\theta_{\mathrm{cl}})\|_2 + 2L\,\|\theta_i-\theta_{\mathrm{cl}}\|_2.
    \end{equation}
\end{lemma}
 
\begin{proof}
    By the triangle inequality and smoothness of both losses,
    $\|\delta g(\theta_i)-\delta g(\theta_{\mathrm{cl}})\|_2
    \le L_{\mathrm{bd}}\|\theta_i-\theta_{\mathrm{cl}}\|_2 + L_{\mathrm{cl}}\|\theta_i-\theta_{\mathrm{cl}}\|_2
    \le 2L\|\theta_i-\theta_{\mathrm{cl}}\|_2$.
\end{proof}
 
\begin{lemma}[MI-Induced Contrastive Separation]\label{lem:mi-sep}
    Let $\{g_r^-\}_{r=1}^R$ be the negative clean gradients from non-anchor subsets. If $\mathcal{L}_{\mathrm{MI}}(G_\Delta)\le\epsilon_{\mathrm{MI}}<\log 2$, then for every $r\in\{1,\ldots,R\}$,
    \begin{equation}\label{eq:mi-sep}
        q_0-q_r \;\ge\; \tau\log\frac{1}{e^{\epsilon_{\mathrm{MI}}}-1} \;>\; 0,
    \end{equation}
    where $q_0=s(g_{\mathrm{bd}},g_{\mathrm{cl}})$ and $q_r=s(g_{\mathrm{bd}},g_r^-)$.
    Furthermore, if at least one negative gradient satisfies $s(g_{\mathrm{bd}}, g_r^-)\ge 0$ (non-degeneracy), then $\kappa = q_0 > 0$.
\end{lemma}
 
\begin{proof}
    By definition, $\mathcal{L}_{\mathrm{MI}}=\log(1+\sum_{r=1}^R\exp((q_r-q_0)/\tau))$. Thus $\mathcal{L}_{\mathrm{MI}}\le\epsilon_{\mathrm{MI}}$ implies $\sum_{r=1}^R\exp((q_r-q_0)/\tau)\le e^{\epsilon_{\mathrm{MI}}}-1$. Since every summand is nonneg., each term satisfies $\exp((q_r-q_0)/\tau)\le e^{\epsilon_{\mathrm{MI}}}-1$. Taking logarithms: $q_0-q_r\ge\tau\log(1/(e^{\epsilon_{\mathrm{MI}}}-1))$. For the second claim, let $r'$ be an index with $q_{r'}\ge 0$. Then $\kappa = q_0 \ge q_{r'} + \tau\log(1/(e^{\epsilon_{\mathrm{MI}}}-1)) > 0$.
\end{proof}
 
\begin{remark}\label{rem:mi-sep-interpretation}
    Lemma~\ref{lem:mi-sep} is stronger than a simple alignment condition $\kappa>0$: it requires the backdoor gradient to be \emph{contrastively} close to the clean-task signal while being \emph{separated} from non-matching clean directions. This encourages $g_{\mathrm{bd}}$ to share the clean-task optimization \emph{structure}---the subspace the GCL mechanism will protect---rather than merely following one transient geometric direction. The non-degeneracy condition ($\exists\, r: s(g_{\mathrm{bd}}, g_r^-)\ge 0$) is mild: negative gradients from random non-anchor subsets of the first-task data span diverse directions, so at least one is expected to have nonnegative cosine similarity with $g_{\mathrm{bd}}$ in practice.
\end{remark}
 
\subsection{InfoNCE as a Lower Bound on Mutual Information}\label{sec:theory-infonce}
 
We establish the theoretical link between the gradient MI loss $\mathcal{L}_{\mathrm{MI}}$ used in Phase~2 and the mutual information $I(g_{\mathrm{bd}};\,g_{\mathrm{clean}})$ between the backdoor and clean optimization signals.
The result below follows the density-ratio argument of van den Oord et al.~\cite{Oord2018RepresentationLW}; we include a self-contained proof for completeness.
 
\begin{proposition}[InfoNCE Lower Bound on Mutual Information]\label{prop:infonce-mi}
    Let $g_{\mathrm{bd}}$ and $g_{\mathrm{clean}}$ be drawn from a joint distribution $p(g_{\mathrm{bd}},g_{\mathrm{clean}})$, and let $\{g_r^-\}_{r=1}^R$ be drawn independently from the marginal $p(g)$.
    For any critic function $f:\mathbb{R}^d\times\mathbb{R}^d\to\mathbb{R}_{>0}$, define
    \begin{equation}\label{eq:infonce-general}
        \mathcal{L}_{\mathrm{InfoNCE}}
        = -\mathbb{E}\!\left[\log\frac{f(g_{\mathrm{bd}},g_{\mathrm{clean}})}{f(g_{\mathrm{bd}},g_{\mathrm{clean}})+\sum_{r=1}^R f(g_{\mathrm{bd}},g_r^-)}\right].
    \end{equation}
    Then
    \begin{equation}\label{eq:mi-bound}
        I(g_{\mathrm{bd}};\,g_{\mathrm{clean}})
        \;\ge\;
        \log(R+1) \;-\; \mathcal{L}_{\mathrm{InfoNCE}}.
    \end{equation}
    In our instantiation, $f(a,b)=\exp(s(a,b)/\tau)$ with $s(a,b)=\frac{\langle a,b\rangle}{\|a\|_2\|b\|_2}$, and the loss reduces to $\mathcal{L}_{\mathrm{MI}}$.
    Consequently, minimizing $\mathcal{L}_{\mathrm{MI}}$ maximizes a lower bound on $I(g_{\mathrm{bd}};\,g_{\mathrm{clean}})$.
\end{proposition}
 
\begin{proof}
    Consider the $(R{+}1)$-way classification task: given $g_{\mathrm{bd}}$ and $R{+}1$ candidate gradient vectors $\{g_0,g_1,\ldots,g_R\}$, exactly one of which (say $g_0=g_{\mathrm{clean}}$) is drawn from the conditional $p(\cdot|g_{\mathrm{bd}})$ and the remaining $R$ are drawn independently from the marginal $p(\cdot)$, classify which is the positive sample.
 
    \textbf{Step 1: Optimal classifier.}
    The posterior probability that the positive sample is at position $j$ is
    \begin{equation*}
        p(j\,|\,g_{\mathrm{bd}},\{g_i\})
        = \frac{p(g_j|g_{\mathrm{bd}})\prod_{i\ne j}p(g_i)}{\sum_{k=0}^R p(g_k|g_{\mathrm{bd}})\prod_{i\ne k}p(g_i)}
        = \frac{p(g_j|g_{\mathrm{bd}})/p(g_j)}{\sum_{k=0}^R p(g_k|g_{\mathrm{bd}})/p(g_k)}.
    \end{equation*}
    The Bayes-optimal classifier thus selects $j^\star=\underset{j}{\arg\max} p(g_j|g_{\mathrm{bd}})/p(g_j)$, and uses the density ratio as its score function.
 
    \textbf{Step 2: Cross-entropy bound.}
    The InfoNCE loss in Eq.~\eqref{eq:infonce-general} is the cross-entropy of a classifier that uses the critic $f$ in place of the true density ratio.
    Since the cross-entropy of \emph{any} classifier is lower-bounded by the entropy of the true posterior, and the Bayes-optimal classifier achieves
    \begin{equation*}
        \mathcal{L}_{\mathrm{InfoNCE}}^{\mathrm{opt}}
        = -\mathbb{E}\!\left[\log\frac{p(g_{\mathrm{clean}}|g_{\mathrm{bd}})/p(g_{\mathrm{clean}})}{\sum_{k=0}^R p(g_k|g_{\mathrm{bd}})/p(g_k)}\right],
    \end{equation*}
    we have $\mathcal{L}_{\mathrm{InfoNCE}}\ge\mathcal{L}_{\mathrm{InfoNCE}}^{\mathrm{opt}}$ for any critic $f$.
 
    \textbf{Step 3: Connecting to MI via the classification interpretation.}

We first define the pointwise density ratio 
$\rho(g) \triangleq p(g\mid g_{\mathrm{bd}})/p(g)$, 
which depends implicitly on $g_{\mathrm{bd}}$.
With the optimal critic $f^\star(a,b) = p(b|a)/p(b)$, the
Bayes-optimal InfoNCE loss evaluates to
\begin{align*}
    \mathcal{L}_{\mathrm{InfoNCE}}^{\mathrm{opt}}
    &= \mathbb{E}\!\left[
        -\log\frac{\rho(g_{\mathrm{clean}})}
        {\rho(g_{\mathrm{clean}}) + \textstyle\sum_{r=1}^{R}\rho(g_r^-)}
      \right] \\
    &= -\,\underbrace{\mathbb{E}[\log\rho(g_{\mathrm{clean}})]}_{
          =\;I(g_{\mathrm{bd}};\,g_{\mathrm{clean}})}
      \;+\;\mathbb{E}\!\left[
        \log\!\Bigl(\rho(g_{\mathrm{clean}})
        + \textstyle\sum_{r=1}^{R}\rho(g_r^-)\Bigr)
      \right].
\end{align*}
Conditioning on $(g_{\mathrm{bd}}, g_{\mathrm{clean}})$ and
applying Jensen's inequality (log concave) over the i.i.d.\
negatives gives
$\mathbb{E}_{\mathrm{neg}}[\log(\rho(g_{\mathrm{clean}}) +
  \sum_r\rho(g_r^-))]
\\  \le \log(\rho(g_{\mathrm{clean}}) + R)$,
since $\mathbb{E}_{p(g)}[\rho(g)] = 1$.
Therefore
\[
    \mathcal{L}_{\mathrm{InfoNCE}}^{\mathrm{opt}}
    \;\le\;
    -\,I(g_{\mathrm{bd}};g_{\mathrm{clean}})
    \;+\;
    \mathbb{E}\!\bigl[\log\bigl(\rho(g_{\mathrm{clean}})+R\bigr)\bigr].
\]
Since $\rho(g_{\mathrm{clean}})\ge 0$, we have
$\log(\rho + R) \le \log(\rho) + \log(1 + R/\rho)$.
Alternatively, noting that
$\rho/(\rho+R) \le 1$ implies
$-\log(\rho/(\rho+R)) \ge 0$, we obtain the weaker but sufficient
bound $\mathcal{L}_{\mathrm{InfoNCE}}^{\mathrm{opt}} \ge
-I + \mathbb{E}[\log(\rho+R)]$.
The standard route to close the bound uses the classification
interpretation: $\mathcal{L}_{\mathrm{InfoNCE}}^{\mathrm{opt}} =
H(C\mid\mathbf{G}) \ge \log(R{+}1) - I(g_{\mathrm{bd}};
g_{\mathrm{clean}})$, which follows from the fact that the
$(R{+}1)$-way Bayes error is controlled by the pairwise
distinguishability between $p(\cdot|g_{\mathrm{bd}})$ and
$p(\cdot)$~\cite{Oord2018RepresentationLW,poole2019variational}.
Rearranging and using $\mathcal{L}_{\mathrm{InfoNCE}} \ge
\mathcal{L}_{\mathrm{InfoNCE}}^{\mathrm{opt}}$ yields
Eq.~\eqref{eq:mi-bound} of Proposition~\ref{prop:infonce-mi}.

\end{proof}
 
\begin{remark}[Tightness and the Role of Negatives]
    The bound becomes tighter as the number of negatives $R$ increases: with more negatives the classification task better approximates the density ratio, so the gap between $\mathcal{L}_{\mathrm{InfoNCE}}$ and the true MI shrinks.
    However, the bound saturates at $\log(R+1)$, so $\mathcal{L}_{\mathrm{MI}}$ can capture at most $\log(R+1)$ nats of MI.
    In practice, moderate $R$ (e.g.\ $R=8$) suffices because the goal is not to estimate MI precisely but to ensure sufficient contrastive separation for the gradient alignment mechanism to operate.
\end{remark}

\subsection{Proof of Theorem~\ref{thm:mi-gap} (MI-Driven Base Gap Minimization)}\label{sec:proof-thm2}
 
\begin{proof}
    \textbf{Part (a): Compatible descent.}
    The directional derivative of $\mathcal{L}_{\mathrm{cl}}$ along $\Delta\theta=-\eta(g_{\mathrm{cl}}+\lambda_{\mathrm{bd}}g_{\mathrm{bd}})$ is
    $\langle g_{\mathrm{cl}},\Delta\theta\rangle=-\eta\|g_{\mathrm{cl}}\|_2^2-\eta\lambda_{\mathrm{bd}}\langle g_{\mathrm{cl}},g_{\mathrm{bd}}\rangle$.
    Since $\kappa>0$, the inner product $\langle g_{\mathrm{cl}},g_{\mathrm{bd}}\rangle=\kappa\|g_{\mathrm{cl}}\|_2\|g_{\mathrm{bd}}\|_2\\>0$, so the expression is strictly negative. Analogously,
    $\langle g_{\mathrm{bd}},\Delta\theta\rangle=-\eta\langle g_{\mathrm{bd}},g_{\mathrm{cl}}\rangle-\eta\lambda_{\mathrm{bd}}\|g_{\mathrm{bd}}\|_2^2<0$.
    Under $L$-smoothness, sufficiently small $\eta$ converts these first-order descent directions into actual loss reductions for both losses.
 
    \textbf{Part (b): Gap reduction.}
    Consider one Phase~3 step. By Taylor expansion of $\Phi$,
    \begin{align*}
        \Phi(\theta+\Delta\theta)-\Phi(\theta)
        &=\langle\delta g(\theta),\,\Delta\theta\rangle+\mathcal{O}(\eta^2)\\
        &=\langle g_{\mathrm{bd}}-g_{\mathrm{cl}},\,-\eta(g_{\mathrm{cl}}+\lambda_{\mathrm{bd}}g_{\mathrm{bd}})\rangle+\mathcal{O}(\eta^2).
    \end{align*}
    Expanding the inner product:
    \begin{align*}
        \langle g_{\mathrm{bd}}-g_{\mathrm{cl}},\,g_{\mathrm{cl}}&+\lambda_{\mathrm{bd}}g_{\mathrm{bd}}\rangle\\
        &=\lambda_{\mathrm{bd}}\|g_{\mathrm{bd}}\|_2^2-\|g_{\mathrm{cl}}\|_2^2+(1-\lambda_{\mathrm{bd}})\kappa\|g_{\mathrm{bd}}\|_2\|g_{\mathrm{cl}}\|_2\\
        &=\Psi(\kappa,\lambda_{\mathrm{bd}}).
    \end{align*}
    Therefore each step yields $\Phi(\theta+\Delta\theta)-\Phi(\theta)=-\eta\Psi+\mathcal{O}(\eta^2)$. When $\Psi>0$, the gap $\Phi$ decreases monotonically. Summing over $E$ steps gives Eq.~\eqref{eq:thm2}. Since $\Phi(\theta_{\mathrm{cl}})>0$ at the start of Phase~3 ($\mathcal{L}_{\mathrm{bd}}$ is large while $\mathcal{L}_{\mathrm{cl}}$ is at its minimum), the bound shows $\Phi(\theta_1^\star)$ is driven toward zero; if it becomes negative, the backdoor loss falls \emph{below} the clean loss, which is even more favorable for persistence.
 
    For the threshold $r^\star$: setting $\Psi=0$ with $r=\|g_{\mathrm{bd}}\|_2/\|g_{\mathrm{cl}}\|_2$ gives the quadratic $\lambda_{\mathrm{bd}}r^2+(1-\lambda_{\mathrm{bd}})\kappa r-1=0$, whose positive root is $r^\star=\bigl[(\lambda_{\mathrm{bd}}-1)\kappa+\sqrt{(\lambda_{\mathrm{bd}}-1)^2\kappa^2+4\lambda_{\mathrm{bd}}}\bigr]/(2\lambda_{\mathrm{bd}})$.
\end{proof}
 
\begin{remark}[Role of MI Alignment in Gap Reduction]
\label{rmk:why-mi}
MI alignment ($\kappa > 0$) contributes to driving $\Phi(\theta_1^\star)$ 
small through two channels, while the initial training state provides the 
margin that makes the mechanism robust.

\smallskip
\noindent\textbf{(i) Enabling joint descent.}
Without $\kappa > 0$, Part~(a) fails: the update 
$-\eta(g_{\mathrm{cl}}+\lambda_{\mathrm{bd}}g_{\mathrm{bd}})$ is not a 
descent direction for both losses simultaneously, so the gap reduction 
framework of Part~(b) does not apply. MI alignment is therefore a 
necessary condition for the entire mechanism.

\smallskip
\noindent\textbf{(ii) Threshold analysis.}
The gradient-ratio threshold $r^\star(\kappa,\lambda_{\mathrm{bd}})$ is 
the positive root of $\Psi=0$, i.e., $\lambda_{\mathrm{bd}}r^2 + 
(1{-}\lambda_{\mathrm{bd}})\kappa\,r - 1 = 0$. At $\kappa=0$, 
$r^\star = 1/\!\sqrt{\lambda_{\mathrm{bd}}}$; at $\kappa=1$, the 
quadratic factors as $(r-1)(\lambda_{\mathrm{bd}}r+1)=0$, giving 
$r^\star=1$ regardless of $\lambda_{\mathrm{bd}}$. The monotonicity 
of $r^\star$ in $\kappa$ depends on the backdoor weight:
\begin{itemize}[leftmargin=*, nosep]
\item When $\lambda_{\mathrm{bd}} < 1$: $r^\star$ \emph{decreases} from 
$1/\!\sqrt{\lambda_{\mathrm{bd}}} > 1$ to $1$ as $\kappa$ grows, directly 
relaxing the condition for $\Psi > 0$.
\item When $\lambda_{\mathrm{bd}} \geq 1$: $r^\star$ \emph{increases} from 
$1/\!\sqrt{\lambda_{\mathrm{bd}}} \leq 1$ to $1$ as $\kappa$ grows. The 
threshold rises but remains at most~$1$.
\end{itemize}
In both regimes, $r^\star \leq \max(1,\,1/\!\sqrt{\lambda_{\mathrm{bd}}})$ 
for all $\kappa \in [0,1]$.

\smallskip
\noindent\textbf{(iii) Effect of $\kappa$ on the reduction rate.}
The sensitivity of $\Psi$ to alignment is 
$\partial\Psi/\partial\kappa = (1{-}\lambda_{\mathrm{bd}})
\|g_{\mathrm{bd}}\|_2\|g_{\mathrm{cl}}\|_2$, which is positive when 
$\lambda_{\mathrm{bd}} < 1$ and negative when $\lambda_{\mathrm{bd}} > 1$. 
Thus, increasing $\kappa$ amplifies $\Psi$ only in the 
$\lambda_{\mathrm{bd}} < 1$ regime. When $\lambda_{\mathrm{bd}} > 1$, 
larger $\kappa$ slightly reduces $\Psi$, but this effect is dominated 
by the $\lambda_{\mathrm{bd}}\|g_{\mathrm{bd}}\|_2^2$ term as long as 
$r \gg 1$.

\smallskip
\noindent\textbf{(iv) Initial condition guarantees $\Psi > 0$.}
Regardless of the regime, $\Psi > 0$ holds at the start of Phase~3 
because $r = \|g_{\mathrm{bd}}\|_2/\|g_{\mathrm{cl}}\|_2 \gg 1$: 
$\mathcal{L}_{\mathrm{bd}}$ is large (the trigger--target association is 
not yet implanted) while $\mathcal{L}_{\mathrm{cl}}$ is at its minimum 
(clean pre-training has converged), so $r$ comfortably exceeds $r^\star$ 
in either case. This large initial margin is the primary reason that 
the gap reduction mechanism operates reliably, independent of the 
specific monotonicity of $r^\star$ or the sign of 
$\partial\Psi/\partial\kappa$.

\smallskip
\noindent In summary, $\kappa > 0$ is necessary to activate the gap 
reduction mechanism~(i), the threshold $r^\star$ remains bounded~(ii), 
the effect of $\kappa$ on $\Psi$ is regime-dependent but secondary~(iii), 
and the initial training state provides the dominant guarantee that 
$\Psi > 0$ throughout Phase~3~(iv).
\end{remark}
 
\subsection{Proof of Theorem~\ref{cor:freerider} (Free-Rider Persistence)}\label{sec:proof-cor}
To prove Theorem~\ref{cor:freerider}, we first need to establish the relation between $\Phi(\theta_t)$ and $\Phi(\theta_1^{\star})$. To this end, we introduce Lemma~\ref{lem:gap-preservation} as below.

\begin{lemma}[Target-Task Loss Gap Preservation]\label{lem:gap-preservation}
    Under the Lipschitz smoothness assumption of the GNN model, after $t{-}1$ benign downstream GCL updates starting from the poisoned checkpoint $\theta_1^\star$, the target-task loss gap at the t-th benign GCL update satisfies
    \begin{equation}\label{eq:thm1}
    \small
    \begin{split}
        \Phi(\theta_t)
        \;\le\;
        \Phi(\theta_1^\star)
        &\;+\;
        \bigl(C_{\mathrm{feat}}
        +C_{\mathrm{str}}\sqrt{m\rho}
        +C_{\mathrm{prop}}\sqrt{\epsilon_{\mathrm{prop}}}
        +C_{\mathrm{label}}\bigr)
        \textstyle\sum_{i=1}^{t-1}\|\Delta\theta_i\|_2\\
        &\;+\;
        L\textstyle\sum_{i=1}^{t-1}\|\Delta\theta_i\|_2^2
        \;+\;
        2L\,D_t\textstyle\sum_{i=1}^{t-1}\|\Delta\theta_i\|_2,
    \end{split}
    \end{equation}
    where $\epsilon_{\mathrm{prop}}=\mathcal{L}_{\mathrm{prop}}(G_\Delta^\star)$ is the propagation loss of the optimized trigger, $m$ and $\rho$ are the size and edge density of the trigger subgraph, $C_{\mathrm{feat}}=C_g C_x B_X$ is the trigger-feature contribution, $C_{\mathrm{str}}=C_g C_A$ is the adjacency contribution, $C_{\mathrm{prop}}$ captures propagation instability, and $C_{\mathrm{label}}=\|\tilde{g}_{\mathrm{cl}}(\theta_{\mathrm{cl}})-g_{\mathrm{cl}}(\theta_{\mathrm{cl}})\|_2$ is the label mismatch between the anchor-set gradient under the target label and the full clean gradient (see Lemma~\ref{lem:mismatch-ref} for exact definitions). $\Delta\theta_i = \theta_{i+1} - \theta_i$ denotes the parameter update at the i-th downstream GCL step. $C_g$, $C_x$, $C_A$ and $B_x$ are defined as the constants bounding the representation perturbation, gradient perturbation and trigger-node features (see Assumption 2 and 4 in Appendix~\ref {sec:theory-assumptions})
\end{lemma}

Lemma~\ref{lem:gap-preservation} separates the per-step drift into four interpretable components:
$C_{\mathrm{prop}}\sqrt{\epsilon_{\mathrm{prop}}}$ bounds the drift from propagation instability ($\mathcal{L}_{\mathrm{prop}}$), $C_{\mathrm{str}}\sqrt{m\rho}$ from the trigger's topological footprint ($\mathcal{R}_{\mathrm{str}}$), $C_{\mathrm{feat}}$ from the trigger-feature magnitude, and $C_{\mathrm{label}}$ from the label mismatch.
The MI objective $\mathcal{L}_{\mathrm{MI}}$ serves a dual role: it minimizes the \emph{initial} gap $\Phi(\theta_1^\star)$ (Theorem~\ref{thm:mi-gap}) and indirectly drives $C_{\mathrm{label}}$ small by aligning the backdoor gradient with the clean-task signal (Lemma~\ref{lem:mi-sep}).

We now prove Lemma~\ref{lem:mi-sep}. 
\begin{proof}
    \textbf{Step~1: Per-step expansion.}
    By $L$-smoothness (Assumption~\ref{ass:smooth}),
    \begin{align}
        \mathcal{L}_{\mathrm{bd}}(\theta_{i+1})
        &\le\mathcal{L}_{\mathrm{bd}}(\theta_i)
        +\langle g_{\mathrm{bd}}(\theta_i),\Delta\theta_i\rangle
        +\tfrac{L}{2}\|\Delta\theta_i\|_2^2,
        \label{eq:bd-smooth}\\
        \mathcal{L}_{\mathrm{cl}}(\theta_{i+1})
        &\ge\mathcal{L}_{\mathrm{cl}}(\theta_i)
        +\langle g_{\mathrm{cl}}(\theta_i),\Delta\theta_i\rangle
        -\tfrac{L}{2}\|\Delta\theta_i\|_2^2.
        \label{eq:cl-smooth}
    \end{align}
    Subtracting Eq.~\eqref{eq:cl-smooth} from Eq.~\eqref{eq:bd-smooth},
    \begin{equation}\label{eq:step-gap}
        \Phi(\theta_{i+1})-\Phi(\theta_i) \;\le\; \langle \delta g(\theta_i),\;\Delta\theta_i\rangle + L\,\|\Delta\theta_i\|_2^2.
    \end{equation}
 
    \textbf{Step~2: Bounding the first-order term.}
    By Cauchy--Schwarz, $\langle \delta g(\theta_i),\Delta\theta_i\rangle\le\|\delta g(\theta_i)\|_2\|\Delta\theta_i\|_2$. By Lemma~\ref{lem:mismatch-traj}, $\|\delta g(\theta_i)\|_2\le\|\delta g(\theta_{\mathrm{cl}})\|_2\\+2L D_t$. By Lemma~\ref{lem:mismatch-ref}, Eq.~\eqref{eq:total-mismatch}, the full gradient mismatch at the reference point satisfies $\|\delta g(\theta_{\mathrm{cl}})\|_2\le C_{\mathrm{feat}}+C_{\mathrm{str}}\sqrt{m\rho}+C_{\mathrm{prop}}\sqrt{\epsilon_{\mathrm{prop}}}+C_{\mathrm{label}}$, where $C_{\mathrm{label}}=\|\tilde{g}_{\mathrm{cl}}(\theta_{\mathrm{cl}})-g_{\mathrm{cl}}(\theta_{\mathrm{cl}})\|_2$ is the label mismatch.

    \textbf{Step~3: Telescoping.}
    Substituting into Eq.~\eqref{eq:step-gap} and summing over $i=1,\ldots,t-1$:
    \begin{align*}
        &\Phi(\theta_t)-\Phi(\theta_1^\star)\\
        &\le \sum_{i=1}^{t-1}\Bigl[\bigl(C_{\mathrm{feat}}+C_{\mathrm{str}}\sqrt{m\rho}+C_{\mathrm{prop}}\sqrt{\epsilon_{\mathrm{prop}}}+C_{\mathrm{label}}+2L D_t\bigr)\|\Delta\theta_i\|_2\\
        &\qquad\qquad+L\|\Delta\theta_i\|_2^2\Bigr],
    \end{align*}
    which is exactly Eq.~\eqref{eq:thm1}.
\end{proof}
 
\begin{remark}[Trajectory Locality]\label{rem:trajectory}
    The trajectory-drift term $2LD_t\sum_i\|\Delta\theta_i\|_2$ arises because the gradient mismatch bound (Lemma~\ref{lem:mismatch-ref}) is anchored at $\theta_{\mathrm{cl}}$. Under effective anti-forgetting (Assumption~\ref{ass:af}), $D_t$ is controlled: stability mechanisms constrain the parameter trajectory to remain in a neighborhood of the historical solution. For replay- and regularization-based methods, $D_t$ grows sublinearly in $t$.
\end{remark}
 
After deriving Lemma~\ref{lem:gap-preservation}, we then demonstrate how to derive Eq~\ref{eq:freerider}. 
By definition, $\mathcal{L}_{\mathrm{bd}}(\theta_t)=\mathcal{L}_{\mathrm{cl}}(\theta_t)+\Phi(\theta_t)$. Applying the anti-forgetting guarantee (Assumption~\ref{ass:af}, Eq.~\eqref{eq:cl-preserve}) to the first term ($\mathcal{L}_{\mathrm{cl}}(\theta_t)$) and Eq.\ref{eq:thm1} of Lemma~\ref{lem:gap-preservation} to the second term ($\Phi(\theta_t)$), we can derive Eq~\ref{eq:freerider} of Theorem~\ref{cor:freerider}:
    \begin{align*}
    \small
        &\mathcal{L}_{\mathrm{bd}}(\theta_t)
        \le \mathcal{L}_{\mathrm{cl}}(\theta_1^\star)+\delta_{\mathrm{cl}}\\
        &+\biggl[\bigl(C_{\mathrm{feat}}+C_{\mathrm{str}}\sqrt{m\rho}+C_{\mathrm{prop}}\sqrt{\epsilon_{\mathrm{prop}}}+C_{\mathrm{label}}+2LD_t\bigr)\sum_{i}\|\Delta\theta_i\|_2\\
        &\qquad+\Phi(\theta_1^\star)+L\sum_{i}\|\Delta\theta_i\|_2^2\biggr].
    \end{align*}
    By Theorem~\ref{thm:mi-gap}, $\Phi(\theta_1^\star)$ is small when $\epsilon_{\mathrm{MI}}$ is small. By Lemma~\ref{lem:mi-sep}, MI alignment also drives $C_{\mathrm{label}}$ small, since the contrastive separation encourages $g_{\mathrm{bd}}$ to align with $g_{\mathrm{cl}}$, reducing the discrepancy between the anchor-set gradient under the target label and the full-dataset clean gradient. The term $\mathcal{L}_{\mathrm{cl}}(\theta_1^\star)+\delta_{\mathrm{cl}}$ is exactly what the GCL system already tolerates for clean-task preservation. Thus the backdoor loss inherits the same protection.
 
\subsection{Structural Stealthiness}\label{sec:theory-stealth}
 
\begin{proposition}[Structural Perturbation Bound]\label{prop:stealth}
    Under Assumptions~\ref{ass:lipschitz-mp} and~\ref{ass:sparse}, the representation perturbation induced by a sparse trigger of $m$ nodes and density $\rho$ satisfies $\max_{v\in\mathcal{A}}\|h_{\Delta,v}^{(K)}-h_v^{(K)}\|_2\le C_x B_X + C_A\sqrt{m\rho}$,
    where $C_x, C_A$ are the Lipschitz constants from Assumption~\ref{ass:lipschitz-mp} and $B_X$ is the trigger-feature bound from Assumption~\ref{ass:sparse}. Both terms decrease with smaller triggers: $C_x B_X$ shrinks with the feature magnitude and $C_A\sqrt{m\rho}$ shrinks with the topological footprint. Moreover, nodes outside the anchor-local $K$-hop computation graph have unchanged representations, making the perturbation \emph{localized}.
\end{proposition}
 
\begin{proof}
    By Eq.~\eqref{eq:lip-mp} and Assumption~\ref{ass:sparse}, $\|r_\Delta^{(K)}(v)\|_2\le C_x B_X+C_A\sqrt{m\rho}$. Thus $\max_{v\in\mathcal{A}}\|r_\Delta^{(K)}(v)\|_2\le C_x B_X + C_A\sqrt{m\rho}$. For the locality claim, the trigger injection $\mathcal{I}(\cdot,G_\Delta)$ only modifies the $K$-hop neighborhood of the anchor node; nodes beyond this range receive unmodified aggregation messages.
\end{proof}
 
\begin{remark}[Joint Stealthiness and Persistence]
    Proposition~\ref{prop:stealth} shows that $\mathcal{R}_{\mathrm{str}}$ serves a dual role. In Theorem~\ref{thm:gap-preservation}, smaller $m\rho$ tightens the gap-preservation bound (persistence). In Proposition~\ref{prop:stealth}, smaller $m\rho$ reduces the visible representation perturbation (stealthiness). The two objectives are thus \emph{synergistic}, not competing.
\end{remark}

\begin{algorithm}[t]
\caption{\sysname{}: Persistent Backdoor Attack on GCL}
\label{alg:overall}
\begin{algorithmic}[1]
\Require First-task training data $\mathcal{D}_1^{\mathrm{tr}}$, target label $y^\star$, poisoning budget $\epsilon$, trigger size $m$, weights $\lambda_{\mathrm{MI}},\lambda_{\mathrm{prop}},\lambda_{\mathrm{s}},\lambda_{\mathrm{bd}}$, MI temperature $\tau$, binarization threshold $\rho_{\mathrm{bin}}$, number of negative gradients $R$, propagation depth $K$, learning rates $\eta_{\Delta},\eta_{\theta}$, trigger optimization steps $E_{\Delta}$, poisoning epochs $E_{\mathrm{poison}}$
\Ensure Optimized trigger $G_{\Delta}^{\star}$ and poisoned checkpoint $\theta_1^{\star}$

\State \textbf{Initialize:} trigger $G_{\Delta}=(X_{\Delta},A_{\Delta})$ with $m$ trigger nodes
\State
\State \textbf{// Phase 1: Clean Pre-training}
\State Train $\theta_{\mathrm{clean}}\leftarrow\arg\min_{\theta}\mathcal{L}_{\mathrm{clean}}(\theta;\mathcal{D}_1^{\mathrm{tr}})$
\State Estimate $g_{\mathrm{clean}}\leftarrow\nabla_{\theta}\mathcal{L}_{\mathrm{clean}}(\theta;\mathcal{D}_1^{\mathrm{tr}})\big|_{\theta=\theta_{\mathrm{clean}}}$
\State
\State \textbf{// Phase 2: Trigger Optimization}
\State Sample anchor set $\mathcal{A}\subset\mathcal{D}_1^{\mathrm{tr}}$ with $|\mathcal{A}|=\epsilon|\mathcal{D}_1^{\mathrm{tr}}|$
\For{$t=1$ to $E_{\Delta}$}
    \State Compute $\mathcal{L}_{\mathrm{bd}}(\theta_{\mathrm{clean}};G_{\Delta},\mathcal{A})$
    \State $g_{\mathrm{bd}}(G_{\Delta})\leftarrow\nabla_{\theta}\mathcal{L}_{\mathrm{bd}}(\theta;G_{\Delta},\mathcal{A})\big|_{\theta=\theta_{\mathrm{clean}}}$
    \State Sample non-anchor clean subsets $\{\mathcal{C}_r^{-}\}_{r=1}^{R}$ from $\mathcal{D}_1^{\mathrm{tr}}\setminus\mathcal{A}$
    \For{$r=1$ to $R$}
        \State $g_r^{-}\leftarrow\nabla_{\theta}\mathcal{L}_{\mathrm{clean}}(\theta;\mathcal{C}_r^{-})\big|_{\theta=\theta_{\mathrm{clean}}}$
    \EndFor
    \State $q_0\leftarrow s(g_{\mathrm{bd}}(G_{\Delta}),g_{\mathrm{clean}})$
    \State $q_r\leftarrow s(g_{\mathrm{bd}}(G_{\Delta}),g_r^{-})$ for $r=1,\ldots,R$
    \State $\mathcal{L}_{\mathrm{MI}}(G_{\Delta})\leftarrow-\log\frac{\exp(q_0/\tau)}{\sum_{r=0}^{R}\exp(q_r/\tau)}$
    \State Construct the anchor-local computation graph $\mathcal{G}_{\mathcal{A}}$
    \State Compute $\mathcal{L}_{\mathrm{prop}}(G_{\Delta})$ from the clean and triggered propagation shifts on $\mathcal{G}_{\mathcal{A}}$
    \State $\mathcal{J}_{\Delta}\leftarrow
    \mathcal{L}_{\mathrm{bd}}(\theta_{\mathrm{clean}};G_{\Delta},\mathcal{A})
    +\lambda_{\mathrm{MI}}\mathcal{L}_{\mathrm{MI}}(G_{\Delta})
    +\lambda_{\mathrm{prop}}\mathcal{L}_{\mathrm{prop}}(G_{\Delta})
    +\lambda_{\mathrm{s}}\mathcal{R}_{\mathrm{str}}(A_{\Delta})$
    \State $G_{\Delta}\leftarrow G_{\Delta}-\eta_{\Delta}\nabla_{G_{\Delta}}\mathcal{J}_{\Delta}$
\EndFor
\State $\hat{A}_{\Delta}\leftarrow\mathbb{I}\{\sigma(A_{\Delta})>\rho_{\mathrm{bin}}\}$
\State $G_{\Delta}^{\star}\leftarrow(X_{\Delta},\hat{A}_{\Delta})$
\State
\State \textbf{// Phase 3: Poisoning Fine-tuning}
\State Initialize $\theta\leftarrow\theta_{\mathrm{clean}}$
\For{$e=1$ to $E_{\mathrm{poison}}$}
    \For{mini-batches $\mathcal{B}_c\subset(\mathcal{D}_1^{\mathrm{tr}}\setminus\mathcal{A})$ and $\mathcal{B}_a\subset\mathcal{A}$}
        \State $\mathcal{L}_{\mathrm{ft}}\leftarrow
        \mathbb{E}_{(x,y)\sim\mathcal{B}_c}\big[\ell(f_{\theta}(x),y)\big]
        +
        \lambda_{\mathrm{bd}}
        \mathbb{E}_{(x,y)\sim\mathcal{B}_a}
        \big[\ell(f_{\theta}(\mathcal{I}(x,G_{\Delta}^{\star})),y^\star)\big]$
        \State $\theta\leftarrow\theta-\eta_{\theta}\nabla_{\theta}\mathcal{L}_{\mathrm{ft}}$
    \EndFor
\EndFor
\State $\theta_1^{\star}\leftarrow\theta$
\State
\Return $\theta_1^{\star},G_{\Delta}^{\star}$
\end{algorithmic}
\end{algorithm}

\section{Algorithm of \sysname{}}
\label{sec:app_alg_overall}

We provide the full pseudocode of \sysname{} in Algorithm~\ref{alg:overall}.
The method consists of three stages: (i) clean pre-training on the first task to obtain a reference model and a clean-task optimization signal, (ii) compatibility- and propagation-aware trigger optimization with frozen model parameters, and (iii) poisoning fine-tuning on the first task.
After this intervention, the attacker exits permanently, and the poisoned checkpoint $\theta_1^\star$ is used to initialize the victim's standard GCL procedure on later benign tasks.

\begin{table*}[t]
\centering
\footnotesize
\setlength{\abovecaptionskip}{3pt}
\setlength{\belowcaptionskip}{0pt}
\setlength{\tabcolsep}{1.5pt}
\renewcommand{\arraystretch}{1.10}
\caption{Overall performance on the \textsc{OGB-Arxiv} and \textsc{OGB-Products} datasets under the task-IL protocol.}
\label{tab:appendix_main_taskIL}
\begin{tabular}{l||cccc||cccc||cccc||cccc||cccc||cccc}
\toprule
Victim & \multicolumn{8}{c||}{\textsc{PDGNNs~\cite{PDGNNs}}} & \multicolumn{8}{c||}{\textsc{CaT-CGL}~\cite{CaT-CGL}} & \multicolumn{8}{c}{\textsc{SSRM}~\cite{SSRM}}\\
\cmidrule(lr){1-1} \cmidrule(lr){2-9} \cmidrule(lr){10-17} \cmidrule(lr){18-25}
\multirow{2.5}{*}{Method}
& \multicolumn{4}{c||}{\textsc{OGB-Arxiv}}
& \multicolumn{4}{c||}{\textsc{OGB-Products}}
& \multicolumn{4}{c||}{\textsc{OGB-Arxiv}}
& \multicolumn{4}{c||}{\textsc{OGB-Products}} & \multicolumn{4}{c||}{\textsc{OGB-Arxiv}}
& \multicolumn{4}{c}{\textsc{OGB-Products}}\\
\cmidrule(lr){2-5}\cmidrule(lr){6-9} \cmidrule(lr){10-13} \cmidrule(lr){14-17} \cmidrule(lr){18-21} \cmidrule(lr){22-25}
& {AA}$\uparrow$ & {AF}$\downarrow$ & {ASR}$\uparrow$ & {BF}$\downarrow$
& {AA}$\uparrow$ & {AF}$\downarrow$ & {ASR}$\uparrow$ & {BF}$\downarrow$
& {AA}$\uparrow$ & {AF}$\downarrow$ & {ASR}$\uparrow$ & {BF}$\downarrow$
& {AA}$\uparrow$ & {AF}$\downarrow$ & {ASR}$\uparrow$ & {BF}$\downarrow$
& {AA}$\uparrow$ & {AF}$\downarrow$ & {ASR}$\uparrow$ & {BF}$\downarrow$
& {AA}$\uparrow$ & {AF}$\downarrow$ & {ASR}$\uparrow$ & {BF}$\downarrow$ \\
\midrule

\textit{w/o} poisoning
& 86.13 & 0.35 & {--} & {--}
& 90.28 & 1.74 & {--} & {--}
& 86.25 & 2.19 & {--} & {--}
& 89.76 & 1.42 & {--} & {--}
& 85.09 & 4.17 & {--} & {--}
& 87.32 & 5.06 & {--} & {--} \\[-1.5ex]

\multicolumn{25}{c}{\tikz \draw[densely dashed] (0,0) -- (1\linewidth,0);} \\

Sub--BD~\cite{sub_backdoor}
& 62.12 & 18.55 & 54.12 & 45.04
& 64.12 & 22.18 & 53.32 & 45.85
& 60.84 & 20.36 & 51.27 & 48.73
& 66.38 & 21.05 & 52.16 & 47.84
& 58.66 & 24.18 & 47.92 & 52.08
& 61.44 & 27.83 & 49.57 & 50.43 \\

GTA~\cite{GTA}
& 63.45 & 17.22 & 51.85 & 47.71
& 65.80 & 19.34 & 49.15 & 50.85
& 64.71 & 16.92 & 55.04 & 44.96
& 63.57 & 23.42 & 48.76 & 51.24
& 62.15 & 21.46 & 52.33 & 47.67
& 59.86 & 29.14 & 45.91 & 54.09 \\

Motif--BD~\cite{motif-backdoor}
& 61.58 & 20.23 & 53.56 & 46.44
& 75.32 & 13.15 & 51.30 & 48.70
& 62.93 & 19.80 & 50.66 & 49.34
& 74.06 & 14.88 & 54.92 & 45.08
& 60.38 & 23.91 & 49.68 & 50.32
& 72.24 & 18.72 & 51.86 & 48.14 \\

UGBA~\cite{UGBA}
& 63.80 & 18.82 & 59.55 & 40.45
& 78.10 & 15.60 & 53.45 & 46.55
& 66.12 & 15.37 & 58.41 & 41.59
& 72.44 & 17.31 & 57.68 & 42.32
& 64.92 & 18.37 & 56.75 & 43.25
& 70.15 & 21.43 & 55.24 & 44.76 \\

NFTA~\cite{NFTA}
& 62.34 & 19.22 & 65.32 & 34.68
& 77.45 & 16.12 & 58.12 & 41.88
& 61.75 & 20.64 & 64.06 & 35.94
& 79.62 & 12.97 & 59.15 & 40.85
& 63.57 & 20.06 & 61.14 & 38.86
& 74.88 & 17.35 & 58.72 & 41.28 \\

GCBA~\cite{GCBA}
& 66.39 & 16.46 & 66.14 & 33.86
& 81.15 & 12.82 & 56.40 & 43.60
& 69.84 & 13.22 & 62.53 & 37.47
& 77.48 & 15.83 & 62.84 & 37.16
& 67.41 & 17.58 & 60.39 & 39.61
& 75.63 & 16.92 & 54.38 & 45.62 \\

CGBA~\cite{CGBA}
& 64.34 & 23.51 & 59.15 & 40.85
& 76.20 & 17.54 & 60.11 & 39.85
& 67.02 & 18.74 & 57.19 & 42.81
& 73.58 & 19.06 & 55.47 & 44.53
& 61.26 & 25.80 & 58.04 & 41.96
& 69.54 & 24.61 & 59.83 & 40.17 \\

SPEAR~\cite{SPEAR}
& 65.78 & 17.79 & 61.32 & 38.68
& 78.35 & 15.48 & 54.02 & 45.95
& 65.46 & 17.85 & 66.73 & 33.27
& 80.41 & 13.34 & 56.92 & 43.08
& 65.10 & 19.75 & 54.67 & 45.33
& 76.82 & 15.76 & 56.12 & 43.88 \\

\midrule

LTB~\cite{latent}
& 71.22 & 12.55 & 78.45 & 21.55
& 77.45 & 11.10 & 78.45 & 21.55
& 73.28 & 9.46 & 81.37 & 18.63
& \underline{81.86} & 8.18 & 79.64 & 20.36
& 70.52 & 12.83 & 76.95 & 23.05
& 80.36 & 11.67 & 78.18 & 21.82 \\

P-Trojan~\cite{P-Trojan}
& \underline{76.84} & \underline{8.92} & \underline{86.73} & \underline{13.27}
& \underline{82.16} & \underline{7.86} & \underline{85.94} & \underline{14.06}
& \underline{77.91} & \underline{6.38} & \underline{89.14} & \underline{10.86}
& 81.27 & \underline{6.04} & \underline{87.68} & \underline{12.32}
& \underline{74.38} & \underline{9.56} & \underline{84.22} & \underline{15.78}
& \underline{82.94} & \underline{8.84} & \underline{85.76} & \underline{14.24} \\

\midrule

\sysname{}
& \textbf{86.47} & \textbf{0.22} & \textbf{99.03} & \textbf{0.73}
& \textbf{89.72} & \textbf{2.08} & \textbf{96.72} & \textbf{3.28}
& \textbf{85.75} & \textbf{2.23} & \textbf{99.76} & \textbf{0.24}
& \textbf{88.86} & \textbf{1.96} & \textbf{95.22} & \textbf{4.78}
& \textbf{84.46} & \textbf{4.17} & \textbf{96.48} & \textbf{2.80}
& \textbf{85.91} & \textbf{5.24} & \textbf{95.73} & \textbf{4.27} \\

\bottomrule
\end{tabular}
\vspace{-2mm}
\end{table*}

\begin{table}[t]
\centering
\footnotesize
\setlength{\abovecaptionskip}{3pt}
\setlength{\belowcaptionskip}{0pt}
\setlength{\tabcolsep}{3.0pt}
\renewcommand{\arraystretch}{1.10}
\caption{Overall performance on the \textsc{OGB-Products} dataset under the class-IL protocol.}
\label{tab:appendix_products_classIL}
\begin{tabular}{l||cccc||cccc}
\toprule
Victim & \multicolumn{4}{c||}{\textsc{PDGNNs}~\cite{PDGNNs}} & \multicolumn{4}{c}{\textsc{CaT-CGL}~\cite{CaT-CGL}}\\
\cmidrule(lr){1-1} \cmidrule(lr){2-5} \cmidrule(lr){6-9}
\multirow{2.3}{*}{Method}
& \multicolumn{4}{c||}{\textsc{OGB-Products}}
& \multicolumn{4}{c}{\textsc{OGB-Products}} \\
\cmidrule(lr){2-5}\cmidrule(lr){6-9}
& {AA}$\uparrow$ & {AF}$\downarrow$ & {ASR}$\uparrow$ & {BF}$\downarrow$
& {AA}$\uparrow$ & {AF}$\downarrow$ & {ASR}$\uparrow$ & {BF}$\downarrow$ \\
\midrule

\textit{w/o} poisoning
& 64.14 & 18.66 & {--} & {--}
& 59.68 & 21.07 & {--} & {--} \\[-1.7ex]

\multicolumn{9}{c}{\tikz \draw[densely dashed] (0,0) -- (0.95\linewidth,0);} \\

Sub--BD~\cite{sub_backdoor}
& 15.44 & 62.45 & 4.21 & 95.79
& 13.72 & 64.08 & 5.08 & 94.92 \\

GTA~\cite{GTA}
& 17.82 & 59.82 & 4.17 & 95.83
& 18.34 & 58.77 & 4.64 & 95.36 \\

Motif--BD~\cite{motif-backdoor}
& 14.21 & 63.11 & 4.25 & 95.21
& 15.95 & 62.16 & 5.37 & 94.63 \\

UGBA~\cite{UGBA}
& 16.96 & 61.41 & 4.16 & 95.84
& 17.22 & 61.04 & 4.91 & 95.09 \\

NFTA~\cite{NFTA}
& 19.45 & 58.15 & 4.78 & 95.22
& 20.18 & 57.34 & 5.81 & 94.19 \\

GCBA~\cite{GCBA}
& 20.87 & 56.40 & 4.75 & 95.25
& 22.46 & 55.82 & 5.66 & 94.34 \\

CGBA~\cite{CGBA}
& 17.21 & 60.86 & 4.75 & 95.25
& 16.74 & 60.29 & 5.92 & 94.08 \\

SPEAR~\cite{SPEAR}
& 18.15 & 59.12 & 4.95 & 95.05
& 19.07 & 58.63 & 6.21 & 93.79 \\

\midrule

LTB~\cite{latent}
& 36.43 & 39.15 & 4.85 & 95.15
& 34.88 & 40.76 & 5.74 & 94.26 \\

P-Trojan~\cite{P-Trojan}
& \underline{46.86} & \underline{33.72} & \underline{67.64} & \underline{32.36}
& \underline{42.15} & \underline{34.38} & \underline{68.92} & \underline{31.08} \\

\midrule

\sysname{}
& \textbf{62.58} & \textbf{19.38} & \textbf{95.31} & \textbf{4.69}
& \textbf{56.82} & \textbf{21.46} & \textbf{94.15} & \textbf{5.85} \\

\bottomrule
\end{tabular}
\vspace{-2mm}
\end{table}

\section{Additional Experimental Details and Results}
\label{sec:appendix_exp}

\subsection{Victim GCL Systems}
\label{app:victim_systems}

We evaluate \sysname{} on four representative GCL victim systems with different anti-forgetting mechanisms: replay-based PDGNNs~\cite{PDGNNs} and CaT-CGL~\cite{CaT-CGL}, regularization-based SSRM~\cite{SSRM}, and parameter-isolation-based TPP~\cite{TPP}.
This selection allows us to examine whether persistent backdoors are tied to a specific victim design or generalize across different ways of preserving historical knowledge.

\noindent \textbf{Replay-based method: PDGNNs.}
PDGNNs is designed for continual learning on expanding graphs, where new node classes and edges arrive sequentially.
A direct graph replay strategy may require storing nodes together with their computation ego-subnetworks, whose memory cost grows rapidly with the GNN receptive field.
PDGNNs addresses this issue through a parameter-decoupled design and Topology-aware Embeddings (TEs), which compress ego-subnetworks into fixed-size replay vectors.
After each task, selected TEs are stored in a Topology-aware Embedding Memory (TEM) and replayed during later training, preserving historical topological information with bounded memory cost~\cite{PDGNNs}.
We include PDGNNs because its explicit replay memory provides a strong setting for testing whether a backdoor can be preserved together with replayed clean knowledge.

\noindent \textbf{Replay-based method: CaT-CGL.}
CaT-CGL is a replay-based continual graph learning framework that constructs compact historical representations for previous tasks and reuses them during subsequent learning~\cite{CaT-CGL}.
Instead of naively replaying all old graph data, it maintains a condensed memory that summarizes previous task information and supports continual training on later tasks.
This makes CaT-CGL complementary to PDGNNs: both rely on replay, but they instantiate historical preservation through different memory constructions.
We include CaT-CGL to test whether \sysname{} remains effective under another replay-style GCL system beyond topology-aware embedding memory.

\noindent \textbf{Regularization-based method: SSRM.}
SSRM studies node-wise graph incremental learning on evolving graphs, where newly arriving tasks can change the graph structure around previously seen nodes~\cite{SSRM}.
Such structural shifts alter the input distribution of old tasks and aggravate catastrophic forgetting.
SSRM mitigates this effect with a regularization-based objective that encourages representations to be less sensitive to structural-shift-induced distribution changes.
Rather than replaying old examples, SSRM preserves historical knowledge by constraining representation changes during graph evolution.
We include SSRM to evaluate \sysname{} beyond replay-based victims and examine whether the backdoor can persist under regularization-based preservation.

\noindent \textbf{Parameter-isolation-based method: TPP.}
TPP is a replay-free graph class-incremental learning method based on task profiling and task-specific graph prompts~\cite{TPP}.
It first estimates the task identity using Laplacian-smoothing-based task prototypes, and then applies the corresponding task-specific prompt with a frozen GNN backbone.
This design effectively isolates task-specific decision behavior into separate prompt components, making the model largely free from conventional parameter overwriting.
We include TPP as a parameter-isolation-style victim to test \sysname{} under a replay-free setting where historical knowledge is preserved through task separation rather than replay or regularization.

\subsection{Details of Evaluation Metrics}
\label{sec:appendix_exp_metrics}

Following the standard protocol in continual graph learning, we evaluate clean continual performance using an \emph{accuracy matrix}.
Given a task sequence of length $T$, we record $M_{\mathrm{acc}}\in\mathbb{R}^{T\times T}$, where $M_{\mathrm{acc}}^{i,j}$ denotes the clean accuracy on task $j$ after the model has finished learning task $i$.
We compute \emph{Average Accuracy} (AA) after learning task $i$ as
\begin{equation*}
\mathrm{AA}_i
=
\frac{1}{i}\sum_{j=1}^{i} M_{\mathrm{acc}}^{i,j},
\qquad i=1,\ldots,T
\qquad (\uparrow).
\end{equation*}
We compute \emph{Average Forgetting} (AF) after learning task $i$ as
\begin{equation*}
\mathrm{AF}_i
=
\frac{1}{i-1}\sum_{j=1}^{i-1}
\left(
M_{\mathrm{acc}}^{j,j}
-
M_{\mathrm{acc}}^{i,j}
\right),
\qquad i=2,\ldots,T
\qquad (\downarrow).
\end{equation*}

For backdoor evaluation, we measure the \emph{Attack Success Rate} (ASR), defined as the proportion of triggered test inputs from the poisoned task that are classified into the attacker-specified target class.
Since the backdoor is injected during task $\mathcal{T}_1$ with target label $y^\star$, we evaluate ASR on $\mathcal{D}^{\mathrm{bd,te}}$ after each continual step:
\begin{equation*}
\mathrm{ASR}_i
=
\mathbb{P}_{(\Delta x,y^\star)\sim\mathcal{D}^{\mathrm{bd,te}}}
\left[
\hat{y}_{\theta_i}(\Delta x)=y^\star
\right],
\qquad i=1,\ldots,T
\qquad (\uparrow).
\end{equation*}

To quantify backdoor persistence, we use \emph{Backdoor Forgetting} (BF), defined as the degradation of ASR from its value after the poisoning task:
\begin{equation*}
\mathrm{BF}_i
=
\mathrm{ASR}_1-\mathrm{ASR}_i,
\qquad i=2,\ldots,T
\qquad (\downarrow).
\end{equation*}

Unless otherwise specified, all reported metrics correspond to their final values after the last task:
$\mathrm{AA}=\mathrm{AA}_T$, $\mathrm{AF}=\mathrm{AF}_T$,
$\mathrm{ASR}=\mathrm{ASR}_T$, and $\mathrm{BF}=\mathrm{BF}_T$.
For AA and ASR, higher is better ($\uparrow$); for AF and BF, lower is better ($\downarrow$).
All values are reported in \%.

\subsection{Baseline Selection and Excluded Methods}
In addition to the evaluated baselines, we note that some continual-learning backdoor attacks, such as CBACL~\cite{CBACL} and the Blind Task Backdoor~\cite{latent}, are not directly comparable under our threat model.
Specifically, these methods assume a more permissive setting in which the adversary can \emph{continuously} intervene during training across multiple subsequent tasks (e.g., injecting poisoned samples, adapting triggers, or manipulating task-specific updates over time).
In contrast, our study focuses on a strictly more restrictive \emph{early-stage-only} attacker who can poison the training data only in the first task and has no access to later tasks.
Evaluating CBACL or Blind Task Backdoor under our protocol would either violate their core assumptions (leading to misleadingly poor results) or require granting additional adversarial capabilities (making the comparison unfair).
Therefore, we exclude these methods and restrict our baselines to approaches that can be instantiated under the same early-stage-only intervention budget.

\subsection{Additional Overall Performance}
\label{sec:appendix_exp_overall}

The additional overall results omitted from the main text are reported in Tables~\ref{tab:appendix_main_taskIL} and~\ref{tab:appendix_products_classIL}.
Table~\ref{tab:appendix_main_taskIL} reports task-IL results on \textsc{OGB-Arxiv} and \textsc{OGB-Products} across PDGNNs, CaT-CGL, and SSRM, while Table~\ref{tab:appendix_products_classIL} reports class-IL results on \textsc{OGB-Products} across PDGNNs and CaT-CGL.
These results are consistent with the main-text observations: \sysname{} achieves the strongest attack effectiveness while maintaining clean utility close to the benign reference, whereas existing graph backdoor attacks and adapted continual-learning backdoor baselines suffer from either severe backdoor forgetting or much larger clean-utility degradation.

For the parameter-isolation-based victim TPP~\cite{TPP}, we observe a different but expected behavior.
Across all evaluated datasets and both GCL protocols, \sysname{} achieves ASR of $100.0\%$ and BF of $0.00\%$.
This is because TPP preserves task-specific knowledge through isolated prompts/parameters rather than shared replay or regularization dynamics.
Once the first-task backdoor is learned by the corresponding task-specific component, later benign tasks do not overwrite it, so the backdoor remains perfectly preserved.
Therefore, TPP further confirms that \sysname{} is effective beyond replay- and regularization-based victims, while its results are reported and discussed separately because its parameter-isolation mechanism makes backdoor persistence almost deterministic.

\subsection{Accuracy-Matrix Analysis of Clean Utility}
\label{sec:appendix_exp_accuracy}

\begin{figure*}[t]
    \centering
    \setlength{\abovecaptionskip}{3pt}
    \setlength{\belowcaptionskip}{0pt}

    \begin{minipage}{0.49\linewidth}
        \centering
        \includegraphics[width=\linewidth]{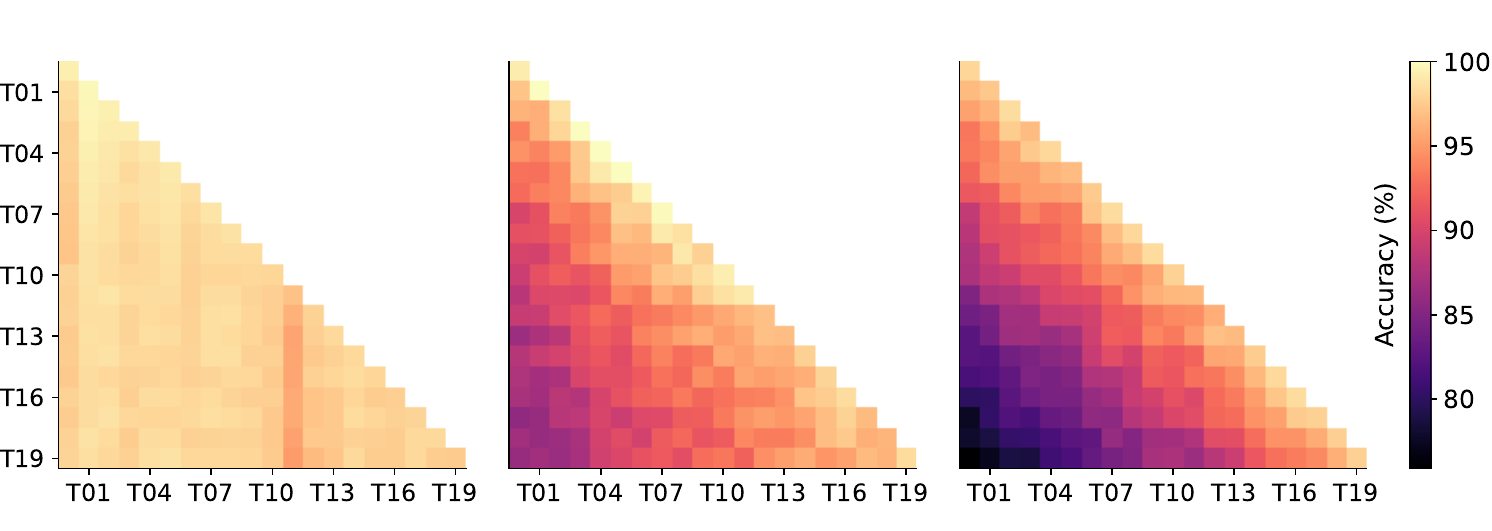}
        \centerline{\small (a) Task-IL}
    \end{minipage}
    \hfill
    \begin{minipage}{0.49\linewidth}
        \centering
        \includegraphics[width=\linewidth]{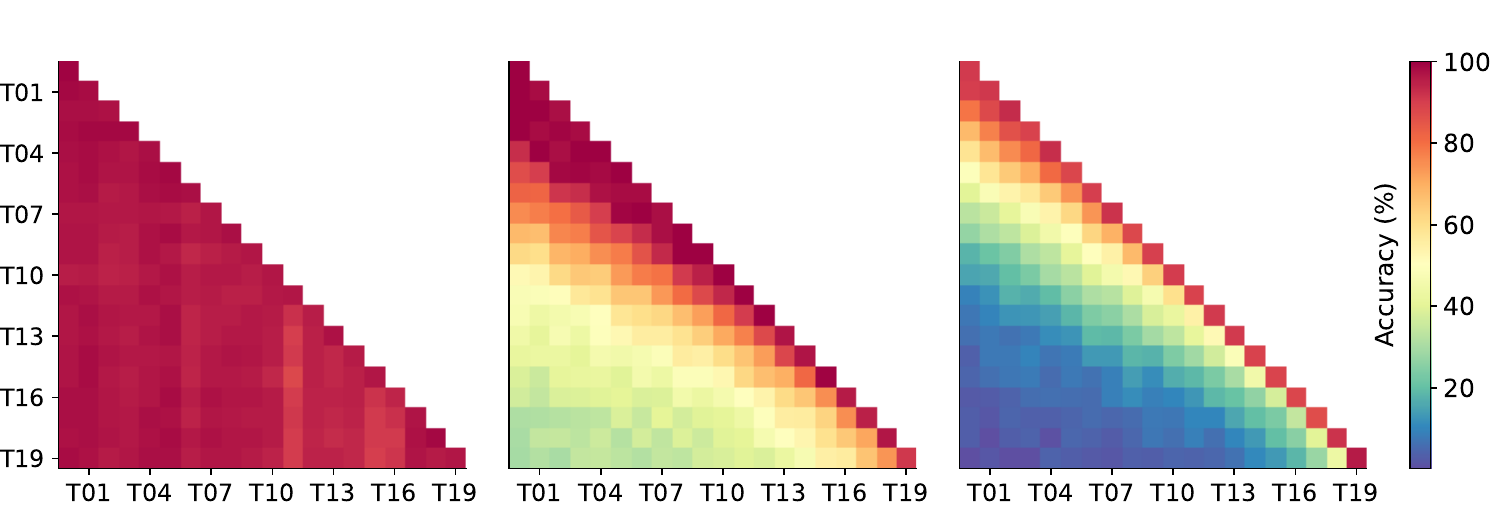}
        \centerline{\small (b) Class-IL}
    \end{minipage}

    \caption{Accuracy matrices of \sysname{} (left), P-Trojan (middle), and GCBA (right) on \textsc{Reddit} under (a) task-IL and (b) class-IL protocols.}
    \label{fig:acc_hotmap}
\end{figure*}

We further analyze clean-utility dynamics using the accuracy matrices of \sysname{}, P-Trojan, and GCBA on the \textsc{Reddit} dataset.
The results are shown in Figures~\ref{fig:acc_hotmap}.
GCBA is selected as the strongest graph backdoor baseline in the main results, while P-Trojan represents the strongest continual-learning backdoor baseline.
Under the task-IL setting, all methods benefit from task identity at inference time, but their forgetting patterns are still different.
GCBA shows a clear degradation on earlier tasks as training proceeds, indicating that its poisoning process damages the victim's ability to preserve historical clean knowledge.
P-Trojan mitigates this issue compared with GCBA, but still exhibits more visible utility degradation than \sysname{}.
In contrast, \sysname{} maintains consistently high accuracy across old and new tasks, showing that the persistent backdoor is not achieved by sacrificing clean-task retention.

The difference becomes more pronounced under the class-IL setting.
GCBA rapidly forgets previously learned tasks, and P-Trojan also suffers clear degradation despite its continual-learning design.
These results explain the poor stealthiness of existing baselines in Table~\ref{tab:main_classIL}: their failures are not only caused by backdoor decay, but also by system-level damage introduced during poisoning.
By contrast, \sysname{} preserves high clean utility across the continual trajectory.
This indicates that optimizing the trigger around the GCL persistence risk in Observation~\ref{obs:persistence_in_gcl} allows the backdoor to remain compatible with continual knowledge preservation while avoiding severe clean-utility collapse.

\subsection{Details of Potential Countermeasures}
\label{sec:appendix_defense_details}

\noindent\textbf{Rationale of defense selection.}
There is no established defense specifically designed for persistent graph backdoors in GCL under our early-stage-only threat model.
Therefore, we select representative defenses from different graph-backdoor defense paradigms to cover the main mitigation strategies explored in prior work.
Specifically, AGNNCert~\cite{Agnncert} represents inference-time certified prediction, AACL~\cite{AACL} represents continual-learning-oriented backdoor mitigation, OD~\cite{OD} represents anomaly-based filtering, RIGBD~\cite{RIGBD} represents robustness-inspired detection, and $\text{A}^2$GBD~\cite{A2GBD} represents adaptive graph purification.
Pruning-based defenses such as PruneG~\cite{PruneG} are covered by this selection because $\text{A}^2$GBD includes edge dropping as one of its mitigation operations.
DShield~\cite{DShield} represents a label-dependent defense paradigm, but it requires target-label information to identify poisoned nodes.
This assumption is incompatible with our GCL setting, where the defender or downstream learner does not retain labeled data from previous tasks for privacy and efficiency reasons.

\vspace{0.4em}
\noindent\textbf{Evaluation protocol.}
All applicable defenses are evaluated on the \textsc{CoraFull} dataset with PDGNNs under both task-IL and class-IL protocols.
For each defense, the attacker poisons only the first task, while all subsequent tasks remain benign.
We report final AA and ASR after the last task.
To isolate the role of structural regularization, we also evaluate \sysname{} \textit{w/o} $\mathcal{R}_{\mathrm{str}}$ under the same defenses.

\vspace{0.4em}
\noindent\textbf{AGNNCert: inference-time certified prediction.}
AGNNCert~\cite{Agnncert} partitions an input graph into multiple subgraphs, runs the victim GNN on each subgraph, and aggregates predictions through majority voting.
We apply it only during evaluation, without modifying training data or model parameters.
As shown in Table~\ref{tab:defense_corafull}, AGNNCert has little effect on either AA or ASR.
This is expected because our backdoor association is already consolidated in the model during the first task; inference-time voting can perturb predictions but cannot erase the learned malicious association.

\vspace{0.4em}
\noindent\textbf{AACL: continual-learning-oriented mitigation.}
AACL~\cite{AACL} introduces correctly labeled decoy-trigger samples to inoculate the model against malicious trigger behavior.
We apply it during first-task training, since poisoning occurs only in the first task, and use the same defensive pattern at evaluation.
AACL moderately reduces ASR but also lowers AA.
This reflects its mechanism: the defense deliberately changes the training and evaluation distribution, which can interfere with graph message passing and normal task features.
The \textit{w/o} $\mathcal{R}_{\mathrm{str}}$ variant is more affected, suggesting that structurally less regularized triggers are more vulnerable to distributional interference.

\vspace{0.4em}
\noindent\textbf{OD: anomaly-based filtering.}
OD~\cite{OD} treats poisoned or triggered nodes as distributional outliers and filters suspicious nodes based on feature or structural deviation.
We apply OD to the poisoned first-task graph before continual training.
OD reduces ASR to some extent while preserving most clean utility.
However, it cannot fully remove \sysname{} because the trigger is optimized not only as a local anomalous pattern, but also as a persistent association compatible with early-task optimization.
The larger ASR reduction for \sysname{} \textit{w/o} $\mathcal{R}_{\mathrm{str}}$ indicates that structural regularization helps make the trigger less distinguishable from benign graph patterns.

\vspace{0.4em}
\noindent\textbf{RIGBD: robustness-inspired detection.}
RIGBD~\cite{RIGBD} uses random edge dropping as a robustness probe: nodes whose predictions fluctuate strongly under edge perturbation are considered suspicious and then mitigated through robust training.
We apply RIGBD to the poisoned first-task graph before continuing the standard GCL pipeline.
RIGBD is stronger than OD, especially against \sysname{} \textit{w/o} $\mathcal{R}_{\mathrm{str}}$.
This is because unregularized triggers rely more heavily on local structural artifacts, making predictions less stable under edge perturbation.
In contrast, \sysname{} retains higher ASR, showing that $\mathcal{R}_{\mathrm{str}}$ reduces structural abnormality and weakens robustness-based detection signals.

\vspace{0.4em}
\noindent\textbf{$\text{A}^2$GBD: adaptive graph purification.}
$\text{A}^2$GBD~\cite{A2GBD} first identifies suspicious nodes and then adaptively applies mitigation operations, including edge dropping, feature smoothing with neighboring node features, and temperature adjustment.
We apply $\text{A}^2$GBD to the poisoned first-task data before continual training.
It yields the largest ASR reduction among the evaluated defenses, but also causes the largest AA loss.
This indicates that strong graph purification can suppress the attack only by aggressively modifying graph structure, node features, or prediction confidence, which also disrupts benign message passing.
The much larger degradation of \sysname{} \textit{w/o} $\mathcal{R}_{\mathrm{str}}$ further confirms that structural regularization is critical for surviving adaptive graph purification.

\vspace{0.4em}
\noindent\textbf{Why $\mathcal{R}_{\mathrm{str}}$ improves defense robustness.}
Across defenses, \sysname{} \textit{w/o} $\mathcal{R}_{\mathrm{str}}$ is consistently less robust, especially against structure-oriented defenses such as RIGBD and $\text{A}^2$GBD.
This trend aligns with the role of $\mathcal{R}_{\mathrm{str}}$: it encourages the trigger subgraph to be sparse and structurally unobtrusive, making the injected pattern statistically and topologically closer to benign connectivity.
By avoiding dense or anomalous edge configurations that purification and robustness-based detection methods typically target, the regularized trigger better survives graph-level sanitization and preserves higher attack effectiveness under defenses.

\end{document}